\documentclass{article}

\usepackage{microtype}
\usepackage{graphicx}
\usepackage{subfigure}
\usepackage{booktabs} % for professional tables
\usepackage[table]{xcolor}
\usepackage{mathtools}
\usepackage{float}
\usepackage{booktabs}
\usepackage[inline]{enumitem}
\usepackage{graphicx}
\usepackage{caption}
\usepackage{amsmath}
\usepackage{thmtools}

\usepackage{cancel}

\usepackage{amsmath,amsfonts,bm}

\def\eqref#1{equation~\ref{#1}}
\DeclareMathAlphabet{\mathsfit}{\encodingdefault}{\sfdefault}{m}{sl}
\SetMathAlphabet{\mathsfit}{bold}{\encodingdefault}{\sfdefault}{bx}{n}

\newcommand{\cL}{\mathcal{L}}
\newcommand{\cM}{\mathcal{M}}
\newcommand{\cN}{\mathcal{N}}

\newcommand{\cP}{\mathcal{P}}

\newcommand{\cT}{\mathcal{T}}

\newcommand*{\triplenorm}[1]{{\left\vert\kern-0.25ex\left\vert\kern-0.25ex\left\vert #1
    \right\vert\kern-0.25ex\right\vert\kern-0.25ex\right\vert}}

\newcommand{\R}{\mathbb{R}}

\renewcommand{\phi}{\varphi}
\newcommand{\eps}{\varepsilon}

\newcommand*{\E}{\mathbb E}

\newcommand*{\defeq}{\coloneqq}
\newcommand*{\rd}{\mathrm{d}}
\newcommand*{\dd}{\, \rd}
\DeclareMathOperator*{\argmin}{argmin}

\newcommand\mf[1]{\mathfrak{#1}}

\usepackage{wrapfig}
\usepackage{url}

\usepackage{tikz}
\usetikzlibrary{arrows,backgrounds,bayesnet,calc,matrix}

\definecolor{linkcolor}{RGB}{74, 102, 146}

\usepackage[
colorlinks=true,allcolors=linkcolor,pageanchor=true,
plainpages=false,pdfpagelabels,bookmarks,bookmarksnumbered,
backref=page,
]{hyperref}
\usepackage[capitalize,noabbrev]{cleveref}

\renewcommand*{\backref}[1]{}
\renewcommand*{\backrefalt}[4]{\ifcase #1 Not cited.%
  \or Cited on page~#2.%
  \else Cited on pages #2.%
  \fi%
}

\usepackage{amsmath, amssymb, amsthm,mathtools,dsfont}
\theoremstyle{plain}
\newtheorem{theorem}{Theorem}[section]
\newtheorem{proposition}[theorem]{Proposition}

\newtheorem{lemma}[theorem]{Lemma}
\newtheorem{corollary}[theorem]{Corollary}
\theoremstyle{definition}

\theoremstyle{remark}

\definecolor{lightpurple}{RGB}{168, 141, 201}

\renewcommand*{\eqref}[1]{(\ref{#1})}

\usepackage{suffix}

\usepackage{hyperref}

\usepackage[accepted]{icml2025}

\usepackage{amsmath}
\usepackage{amssymb}
\usepackage{mathtools}
\usepackage{amsthm}
\usepackage{ulem}
\usepackage[capitalize,noabbrev]{cleveref}

\theoremstyle{plain}
\usepackage[textsize=tiny]{todonotes}

\icmltitlerunning{Wasserstein Flow Matching}
\begin{document}

\twocolumn[
\icmltitle{Wasserstein Flow Matching: Generative
           Modeling Over Families of Distributions}

\icmlsetsymbol{equal}{*}
\icmlsetsymbol{dagger}{$\dagger$}

\begin{icmlauthorlist}
\icmlauthor{Doron Haviv}{equal,WC,MSKCC}
\icmlauthor{Aram-Alexandre Pooladian}{equal,NYU}
\icmlauthor{Dana Pe'er}{MSKCC,HHMI}
\icmlauthor{Brandon Amos}{META,dagger}
\end{icmlauthorlist}

\icmlaffiliation{WC}{Weill Cornell}
\icmlaffiliation{MSKCC}{Memorial Sloan–Kettering Cancer Center}
\icmlaffiliation{HHMI}{Howard Hughes Medical Institute}
\icmlaffiliation{NYU}{Center for Data Science, New York University}
\icmlaffiliation{META}{Meta AI}
\icmlcorrespondingauthor{Doron Haviv}{havivd@mskcc.org}
\icmlcorrespondingauthor{Aram-Alexandre Pooladian}{ap6599@nyu.edu}
\icmlcorrespondingauthor{Brandon Amos}{bda@meta.com}

\icmlkeywords{Machine Learning, ICML}

\vskip 0.3in
]

\printAffiliationsAndNotice{\icmlEqualContribution \icmlMetaFootnote} 

\begin{abstract}
Generative modeling typically concerns transporting a single source distribution to a target distribution via simple probability flows. However, in fields like computer graphics and single-cell genomics, samples themselves can be viewed as distributions, where standard flow matching ignores their inherent geometry. We propose Wasserstein flow matching (WFM), which lifts flow matching onto families of distributions using  the Wasserstein geometry. Notably, WFM is the first algorithm capable of generating distributions in high dimensions, whether represented analytically (as Gaussians) or empirically (as point-clouds). Our theoretical analysis establishes that Wasserstein geodesics constitute proper conditional flows over the space of distributions, making for a valid FM objective. Our algorithm leverages optimal transport theory and the attention mechanism, demonstrating versatility across computational regimes: exploiting closed-form optimal transport paths for Gaussian families, while using entropic estimates on point-clouds for general distributions. WFM successfully generates both 2D \& 3D shapes and high-dimensional cellular microenvironments from spatial transcriptomics data. Code is available at \href{https://github.com/DoronHav/WassersteinFlowMatching/}{Wasserstein Flow Matching}.

\end{abstract}

\section{Introduction}
Today's abundance of data and scalability of training massive neural networks has made it possible to generate hyper-realistic images on the basis of training examples \citep{dalle}, as well as video and audio clips \citep{vyas2023audiobox,xing2023survey}, and, of course, text \citep{bubeck2023sparks}. All of these are instances of generative modeling: given access to finitely many samples from a distribution, devise a scheme which generates new samples from the same distribution. Generative modeling has also been revolutionary in the biomedical sciences, for drug design \citep{jumper2021highly}, and single-cell genomics \citep{lopez2018deep}. Nearly all frameworks exploit the notion that datasets (of, say, genomic profiles of cells, images, videos, or corpora of text documents) are instantiations of probability measures, and the task is to transform a point sampled from random noise to generate a data point that obeys the distribution of interest.\looseness-1

Among the zoo of available generative models, one approach noted for its flexibility and simplicity is Flow Matching (FM) \citep{albergo2022building,lipman2022flow,liu2022flow}. For a fixed target probability measure, FM learns an implicitly defined vector field that can transform a source measure (e.g., the standard Gaussian) to the target.  Unlike discrete time and probabilistic generative models (such as Diffusion Models by \citet{song2020score}), FM learns a deterministic, continuous normalizing flow by regressing onto a simple conditional probability flow. This approach, while originally designed for Euclidean domains, can be readily adopted to Riemannian geometries \citep{chen2023riemannian}. Riemannian flow matching (RFM) is widely used for generating samples over geometries such as spheres, tori, translation/rotation groups, simplices, triangular meshes, mazes, and molecular positions and structures. 

However, modern data can exhibit a richer structure in which the samples themselves are distributions. In computational graphics, 3D models are shape distributions observed through point-clouds. Likewise, recent developments in single-cell genomics analysis have demonstrated that gene-expression profiles from groups of cells aggregated via their mean and covariance can capture cellular microenvironments or highlight fine-grain clusters \citep{haviv2024covariance, persad2023seacells}. For both general distributions and Gaussian settings, it is natural to search for a unified generative model that respects the underlying geometry of the data.

\begin{figure}[t]
  \centering
  \resizebox{\linewidth}{!}{
  \begin{tikzpicture}
    \node[anchor=south west, inner sep=0] (image) at (0,0) {
      \includegraphics[width=3cm]{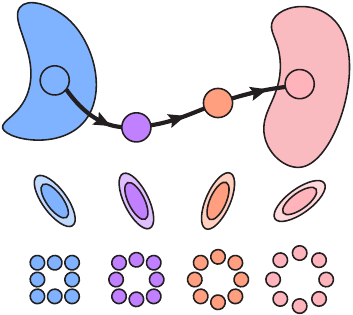}
    };
    \begin{scope}[x={(image.south east)}, y={(image.north west)}]
      \node[anchor=north east] at (0.07, 0.88) {$\mf p_0$};
      \node[anchor=north west] at (1.0, 0.88) {$\mf p_1$};
      \node[anchor=north west] at (0.083, 0.815) {\tiny $\mu$};
      \node[anchor=north west] at (0.78, 0.795) {\tiny $\nu$};
      \node[anchor=north east] at (0.08, 0.48) {\tiny $\cN(m_\mu, \Sigma_\mu)$};
      \node[anchor=north west] at (0.95, 0.48) {\tiny $\cN(m_\nu, \Sigma_\nu)$};
      \node[anchor=north east] at (0.08, 0.24) {\footnotesize $\tfrac{1}{m}\sum_i\delta_{x_i}$};
      \node[anchor=north west] at (0.95, 0.24) {\footnotesize $\tfrac{1}{n}\sum_j\delta_{y_j}$};
    \end{scope}
  \end{tikzpicture}}
  \caption{WFM learns flows between distributions over distributions, where distributions are either analytically represented (Gaussians) or empirically observed through point-clouds.}
   \label{fig:overview}
     \vspace{-4mm}
\end{figure}

\paragraph{Contributions.}
To this end, we introduce \textit{Wasserstein Flow Matching} (WFM), a principled extension of the FM framework lifted to the space of probability distributions. As illustrated in \cref{fig:overview}, a single point in our source and target datasets is itself a distribution. These distributions are either represented analytically as Gaussians or realized through point-cloud samples. Our aim is to learn vector fields acting on the space of probability distributions and match the optimal transport map, which is the geodesic in Wasserstein space. WFM is an instantiation of Riemannian FM ~\citep{chen2023riemannian}, where we train a neural model to learn a continuous normalizing flow (CNF) between distributions over distributions. \looseness-1

We demonstrate the effectiveness of our approach for generative modeling between distributions over Gaussian distributions and distributions over point-clouds, which are realization of general distributions. The former task is motivated by recent directions in single-cell and spatial transcriptomics~\citep{haviv2024covariance, persad2023seacells}, where we consider matching problems over the \textit{Bures--Wasserstein space} (BW), the Gaussian submanifold of the Wasserstein space. In this case, we show that WFM can be further modified, resulting in the Bures--Wasserstein FM (BW-FM) algorithm. We validate BW-FM on a variety of Gaussian-based datasets, where we observe that samples generated by our algorithm are significantly more robust than na\"ive approaches which do not fully exploit the underlying geometry of the data. As an application, we present a generative model for cell states and niches from single-cell genomics data.\looseness-1

\begin{table}[h]
  \centering
  \scalebox{0.8}{
  \begin{tabular}{lccc} \toprule
    \textbf{Method} & \textbf{Data type} & \textbf{Source} & \textbf{Target} \\ \toprule
    FM over $\R^d$ & $x\in\R^d$ & $x\sim p_0$ & $y\sim p_1$ \\
    FM over $\cM$ & $x\in\cM$ & $x\sim \mf p_0$ & $y\sim \mf p_1$ \\
    FM over $\Delta_d$& $\mu\in\cP(\Delta_d)$ & $\mu\sim{\mf p}_0$ & $\nu\sim{\mf p}_1$ \\ \midrule
    Wasserstein FM & $\mu\in\cP(\R^d)$ & $\mu\sim{\mf p}_0$ & $\nu\sim{\mf p}_1$ \\
    $\rightarrow$ Gaussians & $\cN(m, \Sigma)$
    & $\cN(m_\mu,\Sigma_\mu)\sim {\mf p}_0$ & $\cN(m_\nu,\Sigma_\nu)\sim{\mf p}_1$ \\
    $\rightarrow$ Point-Clouds & $\tfrac{1}{n}\sum_i\delta_{x_i}$
    & $\tfrac{1}{m}\sum_{i}\delta_{x_i}\sim{\mf p}_0$ & $\tfrac{1}{n}\sum_{j}\delta_{y_j}\sim{\mf p}_1$ \\
  \end{tabular}}
  \label{tab:overview}
  \vspace{-3mm}
    \caption{Contrasting FM methods over $\R^d$, general manifolds $\cM$, categorical and Dirichlet distributions on the $d$-simplex $\Delta_d$, and finally, our approach, FM problems defined over $\cP(\R^d)$.}
  \vspace{-2 mm}
\end{table}

Generation of general distributions is made possible by two distinct, yet crucial, algorithmic primitives: (1) incorporating transformers in our neural network architecture \citep{vaswani2017attention, lee2019set}, and (2) recent algorithmic advances in entropic optimal transport \citep{pooladian2021entropic}. Our WFM algorithm performs generative modeling in the Wasserstein space, where geodesics are given by pushforwards of optimal transport (OT) maps; see \cref{sec:wass_background} for more information. Since these OT maps lack closed-form solutions for general distributions, we represent distributions as point-clouds and estimate the maps using entropic optimal transport. The permutation equivariance of attention makes transformers a natural basis for our model, inherently capturing the equivariance of Wasserstein geometry while maintaining scalability in high dimensions.

For datasets of distributions in 3D, WFM's performance matches existing generative models. However, due to their reliance on voxelization, current approaches cannot scale to high-dimensional distributions and fail when distributions are realized as point-clouds with variable sizes. Conversely, WFM succeeds in these challenging settings, enabling generative modeling in new domains like synthesizing tissue microenvironments from spatial genomics data. Modeling tissue biology in this generative manner could enhance our understanding of how environment is associated with cell state. In the context of many diseases, most notably cancer and its immune microenvironment, these insights are critical for developing novel therapeutics \citep{binnewies2018understanding}.

\section{Background and related work}

We let $\cP_2(\R^d)$ denote probability distributions over $\R^d$ with finite second moment, and write $\cP_{2,\text{ac}}(\R^d)$ for those with densities. For a probability measure $\mu$ and function $f$, we write $|f|^2_{L^2(\mu)}$ for the squared $L^2(\mu)$ norm. Let $\cM$ be a Riemannian manifold with tangent space $\cT_x\cM$ at $x$. For $x_0 \in \cM$ with velocity $v \in \cT_{x_0}\cM$, the exponential map $\exp_{x_0}(v)$ gives the terminal location, while the logarithmic map $\log_{x_0}(x_1)$ gives the initial velocity for the geodesic from $x_0$ to $x_1$. The set of symmetric (resp. positive definite) matrices over $\R^d$ are denoted by $\mathbb{S}^d$ (resp. $\mathbb{S}^d_{++}$). \looseness-1

\subsection{Riemannian flow matching}\label{sec:rfm}
We first briefly discuss the Riemannian flow matching (RFM) framework of \citet{chen2023riemannian}. Let $\mf p_0$ be the source distribution and $\mf p_1$ be the target distribution over a Riemannian manifold $\cM$, and let $(\gamma_t)_{t\in[0,1]}$ be a curve of probability measures satisfying $\gamma_0 = \mf p_0$ and $\gamma_1 = \mf p_1$. Letting $(w_t)_{t\in[0,1]}$ denote a family of vector fields, we say that the pair $(\gamma_t,w_t)_{t\in[0,1]}$ satisfy the \textit{continuity equation} with respect to the metric $g$, abbreviated to $(\gamma_t,w_t) \in \mf{C}_g$ if
\begin{align}\label{eq:cont_eq}
    \partial_t \gamma_t + \nabla_g \!\cdot\! (\gamma_t w_t) = 0\,,
\end{align}
where $\nabla_{\!g}\cdot$ is the Riemannian divergence operator.

The goal of RFM is to regress a parameterized vector field (e.g., a neural network), written $f_\theta(x, t)\in\cT_{x}\cM$ for $t\in[0,1]$, onto the family $w_t$ by minimizing
\begin{align*}
    \min_\theta \int_0^1 \!\!\int \|f_\theta(z_t,t) - w_t(z_t)\|_{g(z_t)}^2 \dd \gamma_t(z_t) \dd t\,,
\end{align*}
assuming access to a pair $(\gamma_t,w_t)_{t\in[0,1]}$ that satisfies \eqref{eq:cont_eq}. This is not possible in many scenarios. Borrowing insights from recent work (e.g.,~\citet{albergo2022building,lipman2022flow,liu2022flow}), the authors construct a simple vector field that satisfies the continuity equation, resulting in the tractable objective
\begin{align}\label{eq:rfm_obj}
    \min_\theta \int_0^1 \!\!\iint \|f_\theta(x_t,t) - \dot{x}_t\|_{g(x_t)}^2 \dd \mf p_0(x) \dd \mf p_1(y) \dd t\,,
\end{align}
where, for example, $x_t = \exp_{x}((1-t)\log_{x}(y)) \in \cM$, and $\dot{x}_t \in \cT_{x_t}\cM$. For complete discussions and proofs, see \citet[][Section 3.1]{chen2023riemannian}. Once $f_\theta$ is appropriately fit using \eqref{eq:rfm_obj}, we can generate new samples {from $\mf p_1$}:
start by sampling $X_0 \sim \mf p_0$, then follow $\dot{X}_t = f_\theta(X_t,t)$ numerically by discretizing the dynamics given by the exponential map, resulting in $X_1 \sim \mf p_1$. We emphasize that the dynamics are only simulated at inference time and not when training $f_\theta$, commonly known as a \textit{simulation-free} training paradigm.

\subsection{Related work}
\paragraph{Generative models for shapes.}

Paralleling the progress in generative models for natural images, the field of shape generation is rapidly expanding. Many different models have been used from this task, namely generative-adversarial-nets \citep{achlioptas2018learning}, variational autoencoders \citep{gadelha2018multiresolution}, normalizing flows \citep{yang2019pointflow, kim2020softflow, klokov2020discrete}, diffusion \citep{zhou20213d, cai2020learning} and even euclidean FM \citep{wu2023fast}. Thus far, these approaches are limited to uniform-sized samplings of shape distributions in 2D \& 3D, and fail on datasets where realizations are variably sized.

\begin{figure*}[h]
  \centering
  \includegraphics[width=0.9\textwidth]{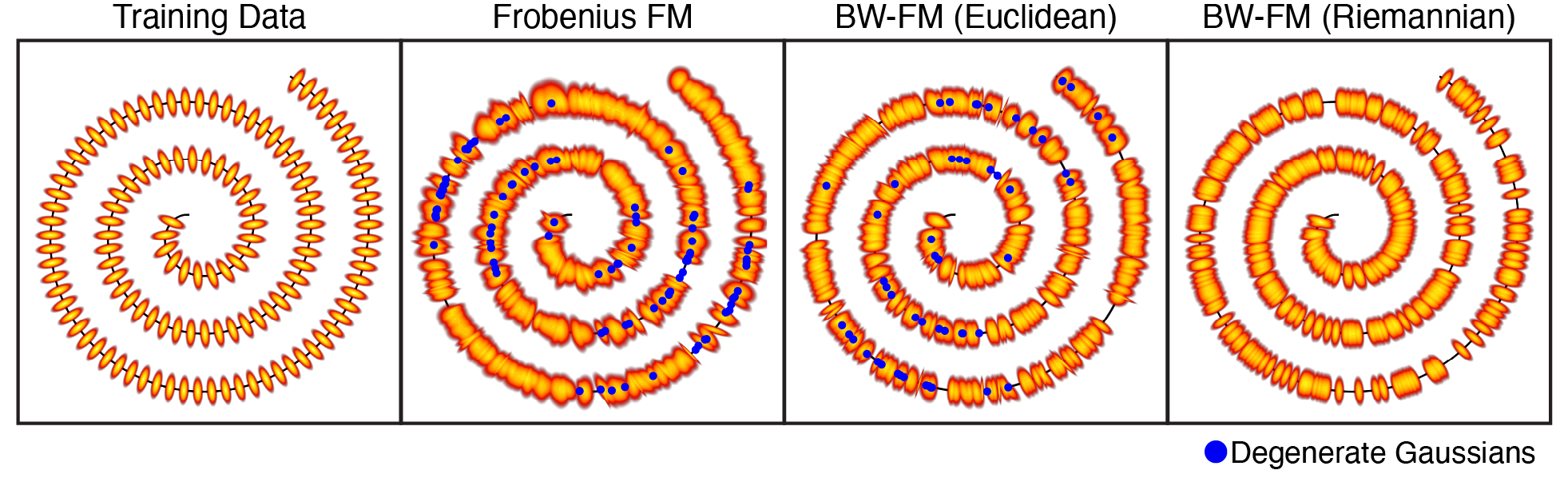}
  \vspace{-4mm}
  \caption{
  In the presence of sufficiently many samples, all methods generate Gaussians along the whole spiral, and our Riemannian BW-FM algorithm produces the most consistent samples. Other methods produce Gaussians with \textit{degenerate} covariance, as they do not model geometry of the data. When there are only few examples, BW-FM accurately reconstructs the training data; see \cref{fig:bw_spirals_8}.
  }\label{fig:bw_spirals}
  \vspace{-4mm}
\end{figure*}

\paragraph{Generative models over families of distributions.}
Our work is not the first to instantiate Riemannian FM with a manifold of probability measures. Two notable works are Fisher FM \citep{davis2024fisher} and
Categorical FM \citep{cheng2024categorical}, which consider the FM algorithm with respect to the Fisher--Rao geometry \cite{amari2016information,
nielsen2020elementary} over the $d$-dimensional simplex $\Delta_d$. The work of \citet{stark2024dirichlet} is similar in spirit, where they focus on the Dirichlet distribution for generation of discrete data. Another related work is that of \citet{atanackovic2024meta}, called Meta FM. Their approach requires pairs of distributions which are already coupled, with the goal of solving FM between a distribution over pairs. In contrast, we emphasize that our proposed Wasserstein FM applies between two separate \textit{uncoupled} distributions over distributions.

\paragraph{Generative models for single-cell genomics.}

Deep learning based generative models have transformed single-cell genomics through various approaches. Variational auto-encoders  \citep{lopez2018deep,  gayoso2022python} have  successfully addressed technical artifacts, integrating multi-modal data and imputing missing features in scRNA-seq data. More recently, Transformer-based foundation models have been noted for their ability to integrate large atlases of data \citep{cui2024scgpt, theodoris2023transfer}. Flow Matching has also emerged as a promising direction \citep{klein2024genotentropicgromovwasserstein, eyring2024unbalancednessneuralmongemaps} for learning both balanced and unbalanced OT maps between cell populations. While these prior FM applications focus on cell-to-cell mappings, our work introduces a new paradigm: generating entire distributions representing cellular populations. This is particularly relevant for spatial genomics, where cellular microenvironments are naturally represented as distributions. WFM enables synthesis of whole cellular neighborhoods, a novel approach in generative modeling for the single-cell field.

\subsection{Wasserstein geometry}\label{sec:wass_background}
The (squared) \textit{$2$-Wasserstein distance} between two probability measures $\mu,\nu \in \cP_{2,\text{ac}}(\R^d)$
  is given by the optimization problem over vector-valued maps $T:\R^d\to\R^d$
\begin{align}\label{eq:wass_dist}
    W_2^2(\mu,\nu) \defeq \min_{T: T_\sharp\mu=\nu} \|{\rm{id}} - T\|_{L^2(\mu)}^2\,,
\end{align}
where the pushforward constraint, written $T_\sharp\mu=\nu$, means that for $X \sim \mu$, $T(X)\sim\nu$. The minimizer to \eqref{eq:wass_dist} is called the \textit{optimal transport (OT) map}, denoted $T^{\mu\to\nu}_\star$ (we abbreviate this to $T_\star$ when clear).
The existence and uniqueness of the optimal transport map under the stated regularity conditions is due to \citet{Bre91}.

The \textit{Wasserstein space} is the space of probability densities with finite second moment endowed with the Wasserstein distance; this space is known to be a metric space \citep{Vil08}. Following the celebrated work of \cite{otto2001geometry}, the Wasserstein space can be formally (meaning, non-rigorously) viewed as a Riemannian manifold, whose properties we now describe in brief; see e.g., \cite{Ambrosio2008} for a rigorous treatment. Following the definition by \citet[Theorem 8.5.1]{Ambrosio2008}, the tangent space at a point $\mu \in \cP_{2,\text{ac}}(\R^d)$ consists of all possible vectors that emanate from $\mu$, written as
\begin{align*}
    \cT_\mu \cP_{2,\text{ac}}(\R^d) \!\defeq\! \overline{\{\lambda(T_\star^{\mu\to\nu} - \mathrm{id})\!:\!\lambda > 0,\nu \in \cP_2(\R^d)\}}^{L^2(\mu)}
\end{align*}
where the overline denotes the closure of the set (i.e., the set and its limit points) in $L^2(\mu)$, and the norm on the tangent space is $L^2(\mu)$. The exponential and logarithmic maps read\looseness-1
\begin{align*}
    v \mapsto \exp_\mu(v) \defeq (\mathrm{id} + v)_\sharp\mu\,, \, \nu \mapsto \log_\mu(\nu) \defeq T_\star^{\mu\to\nu} - \mathrm{id}\,,
\end{align*}
where $\text{id}$ is the identity map. Consequently, the (constant-speed) geodesic, or \textit{McCann interpolation}, between two measures $\mu$ and $\nu$ is given by the curve $(\mu_t)_{t\in[0,1]}$ where
\begin{equation}\label{eq:mccann}
\begin{aligned}
    \mu_t &\defeq (T_t^{\mu\to\nu})_\sharp\mu \defeq ((1-t)\text{id} + tT_\star^{\mu\to\nu})_\sharp\mu \\ &\equiv \exp_\mu(t\log_\mu(\nu))\,,
\end{aligned}
\end{equation}
where the last expression writes the pushforward in terms of the exponential and logarithmic maps. Equivalently, at the level of the random variables, one can write $X_t = (1-t)X_0 + t T_\star^{\mu\to\nu}(X_0)$, where $X_0 \sim \mu$ and $X_t \sim \mu_t$ for any $t \in [0,1]$. Combined with $(v_t)_{t\in[0,1]}$ a suitable family of vector fields, the McCann interpolation satisfies the continuity equation \eqref{eq:cont_eq} over $\R^d$, re-written as
\begin{align}\label{eq:cont_eq_other}
    \partial_t \mu_t + \nabla\cdot(\mu_t v_t) = 0\,, \quad \text{s.t.} \quad \mu_0 = \mu\,, \mu_1 = \nu\,,
\end{align}
where the divergence operator is the usual {Euclidean} one over $\R^d$, thus we write $(\mu_t,v_t) \in \mf{C}$. The link between the constant speed geodesics and the $2$-Wasserstein distance can be viewed from the celebrated Benamou--Brenier formulation of optimal transport \citep{benamou2000computational}:
\begin{align}\label{eq:bb_formulation}
    W_2^2(\mu,\nu) = \inf_{(\mu_t,v_t) \in \mathfrak{C}} \int_0^1 \|v_t\|^2_{L^2(\mu_t)} \dd t\,.
\end{align}
The optimal curve of measures is given by the constant-speed geodesics described above, and the optimal velocity field is given by
\begin{align}\label{eq:vectorfield}
    v_t = (T_\star^{\mu\to\nu} - \text{id})\circ(T_t^{\mu\to\nu})^{-1}\,.
\end{align}
The vector field \eqref{eq:vectorfield} should be interpreted as the time-derivative of the McCann interpolation, with $X_0 \sim \mu$
\begin{align*}
    \dot{X}_t &= (T_\star^{\mu\to\nu} - \text{id})(X_0) \\&= (T_\star^{\mu\to\nu} - \text{id})\circ(T_t^{\mu\to\nu})^{-1}(X_t)\,.
\end{align*}

\subsubsection{Bures--Wasserstein (BW) space}\label{sec:bw_background}
A known special case of the Wasserstein space is the \textit{Bures--Wasserstein} space, which consists of the submanifold of non-degenerate Gaussians parameterized by means and covariances $\{(m,\Sigma) : m \in \R^d, \Sigma \in \mathbb{S}^d_{++}\}$, endowed with the Wasserstein metric. We provide a brief exposition on the geometry of the Bures--Wasserstein space and refer to~\citet{lambert2022variational} for detailed calculations and explanations, as we follow their notation conventions.\looseness-1

The OT map between $\cN(m_\mu,\Sigma_\mu)$ and $\cN(m_\nu,\Sigma_\nu)$ has a closed-form expression \citep{gelbrich1990formula}:\looseness-1
\begin{align*}
    T_{\star}(x) &\defeq m_\nu + C^{\mu\to \nu}(x-m_\mu) \\ &\defeq m_\nu + \Sigma_\mu^{-\tfrac12}(\Sigma_\mu^{\tfrac12}\Sigma_\nu\Sigma_\mu^{\tfrac12})^{\tfrac12}\Sigma_\mu^{-\tfrac12}(x-m_\mu)\,.
\end{align*}
As this map is affine, it is clear that the McCann interpolation between two Gaussians is always Gaussian. More generally, we have the succinct representation of the tangent space at a point in the Bures--Wasserstein space
\begin{align*}
    \cT_\mu{\rm{BW}}(\R^d) \defeq \{a + S({\rm{id}} - m_\mu) \, : \, a \in \R^d, S \in \mathbb{S}^d \}\,,
\end{align*}
and the tangent space norm at $\mu$ can be written as:
\begin{align*}
    \|(a,S)\|^2_{{\rm{BW}}(\mu)} \defeq \|a\|^2 + {\rm {Tr}}(S^2 \Sigma_\mu)
\end{align*}
With the above, it is straightforward to compute the McCann interpolation $\mu_t = (T_t)_\sharp\mu = \cN(m_t,\Sigma_t)$, with
\begin{equation}\label{eq:mccann_cov} 
\begin{aligned}
    m_t &\defeq (1-t)m_\mu + tm_\nu \,, \\ 
    \Sigma_t &\defeq T_tAT_t, \, T_t = (1-t)I + tC^{A\to B} \,
\end{aligned}
\end{equation}
We can relate the Euclidean and Riemannian time-derivatives of $\Sigma_t$ through the following manipulation:
\begin{align}\label{eq:sigt_dot_euclidean}
    \dot{\Sigma}_t^{\rm E} = \dot{T}_t A T_t + T_t A\dot{T}_t = \dot{\Sigma}_t^{\rm{BW}} \Sigma_t + \Sigma_t \dot{\Sigma}_t^{\rm{BW}}\,.
\end{align}
To this end, we can draw parallels to \eqref{eq:vectorfield} by writing
\begin{equation}\label{eq:sigt_dot}
\begin{aligned}
    &\dot{m}_t = m_\nu - m_\mu\,,\\ 
    &\dot{\Sigma}_t^{ \rm{BW} } = (C^{A\to B} - I)((1-t)I + tC^{A\to B})^{-1}\,.
\end{aligned}
\end{equation}

\section{Flow matching over the Wasserstein space}\label{sec:main_wfm}

\begin{figure*}[t]
  \centering
  \includegraphics[width=\textwidth]{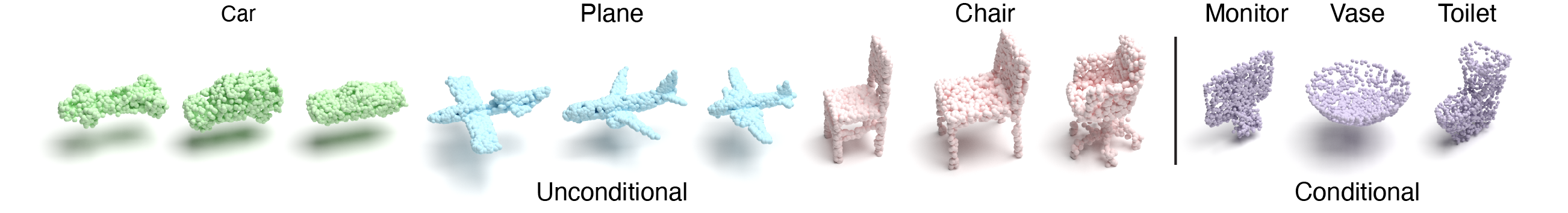}
  \vspace{-8mm}
  \caption{\textit{{Left.}} Synthesized samples from WFM trained on the \textit{cars}, \textit{planes} or \textit{chairs} datasets. \textit{{Right.}} Examples generated conditionally from the \textit{same initial noise} via a WFM model trained on the complete 40-class ModelNet dataset.}\label{fig:wfm_shapenet}
  \vspace{-5mm}
\end{figure*}

Let $\mf p_0$ and $\mf p_1$ denote probability measures over the Wasserstein space. Our goal is to learn a vector field that transports the family of measures $\mathfrak{p}_0$ to the family $\mathfrak{p}_1$. WFM learns to map source to target by regressing onto Wasserstein geodesics between samples $\mu \sim \mathfrak{p}_0$ and $\nu \sim \mathfrak{p}_1$.\footnote{Crucially, we are \textit{not} learning the OT map between $\mathfrak{p}_0$ and $\mathfrak{p}_1$.} To accomplish this, we pass in the McCann interpolation $\mu_t$ and optimal velocity field $v_t$ in the Riemannian FM objective \eqref{eq:rfm_obj}, resulting in our Wasserstein FM (WFM) objective:
\begin{align}\label{eq:fm_wass}
    \min_\theta \int_0^1 \!\!\iint \|f_\theta^{\texttt{geo}}(\mu_t,t) - v_t\|_{L^2(\mu_t)}^2 \dd\mathfrak{p}_0\dd\mathfrak{p}_1 \dd t\,.
\end{align}

See \Cref{sec:wfm_derivation} for a derivation of this training objective. It remains to ensure that this objective describes valid conditional probability flows between families of distributions.

First and foremost, we require two levels of regularity on the distributions of distributions $\mathfrak{p}_0$ and $\mathfrak{p}_1$: 

\begin{enumerate}
\vspace{-2mm}
\item \textbf{Outer continuity:} The source measure $\mathfrak{p}_0$ satisfies smoothness conditions (as in Appendix A of \citet{chen2024probabilistic}) ensuring the conditional paths $\mathfrak p_t(\cdot|\nu)$ can be marginalized over $\mathfrak{p}_1$ to form a valid flow
\vspace{-2mm}
\item \textbf{Inner continuity:} For any $\mu \sim \mathfrak{p}_0$ and $\nu \sim \mathfrak{p}_1$, there exists an optimal transport map between $\mu$ and $\nu$ making the conditional flows well-defined.
\vspace{-2mm}
\end{enumerate}

The condition of ``inner continuity" is fairly mild, as this is ensured for any distribution $\mu \sim \mathfrak p_0$ with density. For Gaussian distributions, inner continuity holds naturally. For general distributions, we assume continuity but work with point-clouds as empirical realizations to approximate OT maps with statistical guarantees (see \cref{sec:eot_stat}) and computational efficiency~\citep{flamary2021pot,cuturi2022optimal}. There exists a slight gap between our theoretical results which consider classic OT and our practical implementation which uses entropy-regularized OT approximations. This aligns with other geometric methods (see Algorithm 1 by \citet{chen2023riemannian}) that rely on numerical approximations when exact solutions are not tractable. The ``outer continuity" condition is purely technical, and it serves the same role as in prior work. Our training algorithm is described in~\cref{alg:WFM}, and \cref{supp_sec:training} contains precise details on our neural network design.

Finally, we prove that regressing onto vector fields results in valid conditional probability flows between distributions of distributions; see \cref{sec:cond_flow} for a proof.

\begin{proposition}\label{prop:cond_prob_main}
Conditional probability paths $\mathfrak p_t(\cdot|\nu)$ generated by conditional vector fields of the form $v_t$ (recall \eqref{eq:vectorfield}) satisfy $\mathfrak p_1(\cdot|\nu) = \nu$.
\end{proposition}
\vspace{-2mm}
To prove this result, we following a general recipe by \citet{chen2023riemannian} with calculations specific to the Wasserstein space. First, we prove that the Wasserstein distance is a \textit{pre-metric} (see Lemma \ref{lem:wasserstein_metric}). Then, we prove that the conditional vector fields $v_t$ are defined as a function of this pre-metric (see Lemma \ref{lem:metric_to_vel}). With these calculations, the proof of Proposition \ref{prop:cond_prob_main} then follows immediately from Theorem 3.1 of \citet{chen2023riemannian}.

\begin{figure*}[t]
\begin{minipage}[t]{0.48\textwidth}
\vspace{0pt}  
\begin{algorithm}[H]
\caption{Wasserstein FM Training}
\label{alg:WFM}
\begin{algorithmic}
\STATE {\bfseries Data:} base $\mathfrak{p}_0 \in \mathcal{P}(\mathcal{P}(\mathbb{R}^d))$, target $\mathfrak{p}_1 \in \mathcal{P}(\mathcal{P}(\mathbb{R}^d))$, $\text{geo} \in \{\text{BW},\text{General}\}$
\STATE {\bfseries Init:} Parameters $\theta$ of $f_\theta^{\text{geo}}$
\REPEAT
\STATE Sample time $t \sim \mathcal{U}(0,1)$
\STATE Sample source measure $\mu \sim \mathfrak{p}_0$
\STATE Sample target measure $\nu \sim \mathfrak{p}_1$
\IF{geo is BW}
\STATE $\mu_t \leftarrow (m_t,\Sigma_t)$ via equation 8
\STATE $v_t \leftarrow (\dot{m}_t,\dot{\Sigma}_t^{\text{BW}})$ via equation 9
\ELSE
\STATE $\mu_t \leftarrow$ Approximate via equation 4
\STATE $v_t \leftarrow$ Approximate via equation 7
\ENDIF
\STATE $\ell(\theta) \leftarrow \|f_\theta^{\text{geo}}(\mu_t,t) - v_t\|^2_{L^2(\mu_t)}$
\STATE $\theta \leftarrow \text{optimizer\_step}(\theta, \ell(\theta), \nabla_\theta \ell(\theta))$
\UNTIL{Convergence}
\end{algorithmic}
\end{algorithm}
\end{minipage}%
\hfill%
\begin{minipage}[t]{0.48\textwidth}
\vspace{0pt}  
\begin{algorithm}[H]
\caption{BW$(\mathbb{R}^d)$ Generation}
\label{alg:bwsim}
\begin{algorithmic}
\STATE {\bfseries Data:} Trained $f_\theta^{\text{BW}}$, step size $h = 1/N$
\STATE {\bfseries Init:} $\mathcal{N}(m_0, \Sigma_0) \sim \mathfrak{p}_0$
\FOR{$k=0,\ldots,N-1$}
\STATE $(s_k, S_k) \leftarrow f_\theta^{\text{BW}}((m_{kh},\Sigma_{kh}), kh)$
\STATE $m_{(k+1)h} \leftarrow m_k + hs_k$, $U_k \leftarrow (I + hS_k)$
\STATE $\Sigma_{(k+1)h} \leftarrow U_k\Sigma_{kh}U_k$
\ENDFOR
\STATE {\bfseries Return:} $\mathcal{N}(m_{Nh},\Sigma_{Nh})$
\end{algorithmic}
\end{algorithm}
\vspace{-4.5mm}  
\begin{algorithm}[H]
\caption{General Distribution Generation}
\label{alg:pointcloud_sim}
\begin{algorithmic}
\STATE {\bfseries Data:} Trained $f_\theta^{\text{PC}}$, step size $h = 1/N$
\STATE {\bfseries Init:} $\mu_0 \sim \mathfrak{p}_0$, $\hat{\mathbf{X}}_0 = \{X_1,\ldots,X_n\} \sim \mu_0$
\FOR{$k=0,\ldots,N-1$}
\STATE $\hat{\mathbf{X}}_{(k+1)h} \leftarrow \hat{\mathbf{X}}_{kh} + hf_\theta^{\text{PC}}(\hat{\mathbf{X}}_{kh}, kh)$
\ENDFOR
\STATE {\bfseries Return:} $\hat{\mathbf{X}}_{Nh}$
\end{algorithmic}
\end{algorithm}
\end{minipage}
\vspace{-5mm}
\end{figure*}

\subsection{WFM over the Bures--Wasserstein space}
First suppose $\mf p_0$ and $\mf p_1$ are distributions over Gaussians, meaning that a batch of samples drawn from $\mf p_0$ and $\mf p_1$ consists of a batch of mean-covariance pairs. Here, the dynamics are straightforward: the interpolant is the McCann interpolation, and the velocity field over the Bures--Wasserstein manifold is also known (see \eqref{eq:mccann_cov} \& \eqref{eq:sigt_dot}, respectively). Since $\mu_t$ is $(m_t,\Sigma_t)$, the neural network is parameterized as $f_\theta^{\text{BW}} : \R^d \times \mathbb{S}^d_{++} \to \R^d \times \mathbb{S}^d$, and the norm on the tangent space simplifies the computations considerably. Our final training objective becomes
\begin{align*}
    \int_0^1 \!\!\iint \|f_\theta^{\text{BW}}((m_t,\Sigma_t),t) - (\dot{m}_t,\dot{\Sigma}_t^{\rm{BW}})\|_{{\rm{BW}}}^2 \dd\frak{p}_0\dd\frak{p}_1\dd t\,.
\end{align*}

\subsection{WFM over distributions of general distributions}

In the case of general distributions, we lose closed-form interpolations. However, we can hope to proceed so long as we have an \textit{approximation} of the optimal transport map between them, written $\hat{T}$, based on empirical samples (point-clouds) drawn from the distributions. There are many works on the approximation of these maps on the basis of samples; see \citet{hutter2021minimax,divol2022optimal,manole2021plugin,pooladian2021entropic}. 

While existing point-cloud generative models work with uniform-sized samples from high-resolution CAD models, many datasets provide inherently variable-sized samples. Though we assume these samples come from underlying continuous distributions, we only observe finite, discrete realizations. This natural variation in sample sizes necessitates our development of WFM to handle point-clouds of non-uniform size, making WFM applicable across both high-resolution shape modeling and naturally discrete datasets.

We approximate optimal transport maps using entropic OT \cite{cuturi2013sinkhorn}, which is computationally efficient on GPUs via Sinkhorn's algorithm \citep{sinkhorn1964relationship}. Let $\bm{X}$ and $\bm{Y}$ represent point-clouds described by data from $\mu \sim \mf p_0$ and $\nu\sim\mf p_1$ respectively, and let $\hat{T}^{\mu\to\nu}$ denote the \textit{entropic} approximation of the optimal transport map (see \cref{app:eot} for full details). The objective \eqref{eq:fm_wass} can be approximated by
\begin{align*}
\int_0^1 \!\!\iint\|f_\theta^{\text{PC}}(\hat{\bm{X}}_t,t)- ((\hat{T}^{\mu\to\nu}(\bm{X})-\bm{X}))\|_2^2 \dd\frak{p}_0\dd\frak{p}_1 \dd t\,,
\end{align*}
where $\hat{\bm{X}}_t = (1-t)\bm{X} + t\hat{T}^{\mu\to\nu}(\bm{X})$ represents a discretized McCann interpolation. We parameterize $f_{\theta}^{PC}$ with a transformer, leveraging the natural alignment between the permutation equivariance of both OT maps and self-attention. Unlike point-cloud models based on voxelization, transformers maintain efficiency in high dimensions.
\vspace{-2mm}

\subsection{Generation}
Once $f_\theta^{\texttt{geo}}$ is trained, we can generate new samples as in Riemannian FM. For the Bures--Wasserstein space, we appeal to the closed-form exponential and logarithmic maps; see \cref{alg:bwsim}. We emphasize that the appropriate Riemannian updates are crucial to obtain non-degenerate final samples. For general distributions, we perform a standard Euler discretization of the learned flow on their point-cloud realizations; see \cref{alg:pointcloud_sim}.

\section{Results}\label{sec:exp}
\subsection{Families of Gaussians}\label{subsec:exp_bw_fm}
We first demonstrate our flow matching framework between measures of Gaussians on synthetic and real datasets. For comparable baselines in each scenario, we construct two simpler flow matching approaches for Gaussian generation: (1)
\textit{Frobenius FM}, which concatenates the mean and covariances, and trains on $\dot{\Sigma}_t^{{\rm{E}}}$ (equation \eqref{eq:sigt_dot_euclidean}) with respect to the squared-Frobenius norm, and (2) BW-FM (Euclidean), which tries to match $\dot{\Sigma}^{{\rm E}}_{t}$ but still under the BW geometry.

\begin{table}[h]
\vspace{-1mm}
\scalebox{0.8}{
\begin{tabular}{l|c|c|c}
& BW-FM (R) & BW-FM (E) & Frobenius FM \\
\hline
Spiral - 16 (2D) & $\mathbf{2.98 \cdot 10^{-4}}$ & $4.00 \cdot 10^{-4}$ &  $1.03 \cdot 10^{-3}$ \\
Spirals - 128 (2D) & $\mathbf{1.28 \cdot 10^{-3}}$ & $1.70 \cdot 10^{-3}$ &  $2.69 \cdot 10^{-3}$ \\ 
Two Moons (2D) & $\mathbf{1.84 \cdot 10^{-4}}$ &  $8.96 \cdot 10^{-4}$ &  $1.30 \cdot 10^{-3}$ \\ 
Sphere (3D) & $\mathbf{6.65 \cdot 10^{-4}}$ &  $2.14 \cdot 10^{-3}$ &  $2.25 \cdot 10^{-3}$ \\ 
 \hline
 \hline
Cities (2D) & $\mathbf{1.88 \cdot 10^{-4}}$ &  $7.26 \cdot 10^{-3}$ &  $1.75 \cdot 10^{-3}$ \\ 
ECG (15D) & $\mathbf{9.24 \cdot 10^{-2}}$ &  $3.26 \cdot 10^{-1}$ &  $3.98 \cdot 10^{-1}$ \\ 
seqFISH (16D) & $\mathbf{3.11\cdot10^{-1}}$ &  $5.31\cdot10^{-1}$ &  $6.82\cdot10^{-1}$ \\ 
scRNA-seq (32D) & $\mathbf{1.31}$ &  $2.74$ &  $3.21$ \\ 
\hline
\end{tabular}}
\vspace{-2mm}
\caption{
{Average min. $W^{2}_2$ distance between each generated Gaussian and the reference datasets. Despite identical training schemes, BW-FM (R) outperforms other approaches on both synthetic and real data across several dimensions.}
}
\label{tab:bw_metrics}
\vspace{-2mm}
\end{table}

In all cases, we assess the quality of the learned flows by computing the the minimum distance between each generated Gaussian and the dataset using the (squared) $2$-Wasserstein distance; see \cref{tab:bw_metrics}. Notably, the flows generated by Frobenius FM and BW-FM (Euclidean) do not strictly adhere to the geometry of the Bures-Wasserstein manifold, requiring synthesized covariance matrices to be projected onto the space of PSD matrices. In contrast, the Riemannian BW-FM algorithm consistently produces valid and accurate results across all dimensions and datasets.

\subsubsection{Toy Datasets}\label{subsubsec:bw_toy}

As a first test, we design a dataset of Gaussians centered on a spiral; See Figures \ref{fig:bw_spirals} \& \ref{fig:bw_spirals_8}. When there were only few samples, only Riemannian BW-FM reconstructed the data, despite other benchmark methods following identical training regimes. On the complete 128-sample dataset, generalizes beyond the training data and synthesizes novel Gaussians lying on the spiral. BW-FM shares this generalization feature with standard FM, and is able to learn the structure underlying the measure from the training data.

\subsubsection{Single-Cell Genomics}\label{subsubsec:bw_real}
Spatial transcriptomics are a set of techniques which build on single-cell genomics and preserve physical information of cells' location in tissues and assay their gene expression. \citet{haviv2024covariance} demonstrated that a cell's microenvironment can be effectively characterized using the mean and covariance of the surrounding cells gene expression. This representation captures key features of cellular neighborhoods and transforms spatial transcriptomics datasets into a measure within the Bures--Wasserstein space, highlighting the value of generative modeling in this context.

During embryogenesis, specific regions within the primitive gut tube differentiate into organs such as the liver or lungs based on interactions between the gut and surrounding mesenchyme \citep{nowotschin2019emergent}. Applied on environments of gut-tube cells from a seqFISH dataset of mouse embryogenesis \citep{lohoff2022integration}, BW-FM synthesized Gaussian cellular niches conditioned on organ labels, thus demanding an understanding of the interplay between spatial context and phenotype. Despite the intricate nature of the gastrulation process, compounded by the dataset's dimensionality, BW-FM can accurately generate organ-specific niches; see \cref{fig:bw_gut_supp}.

\begin{figure}[h]
  \centering
  \includegraphics[width=0.45\textwidth]{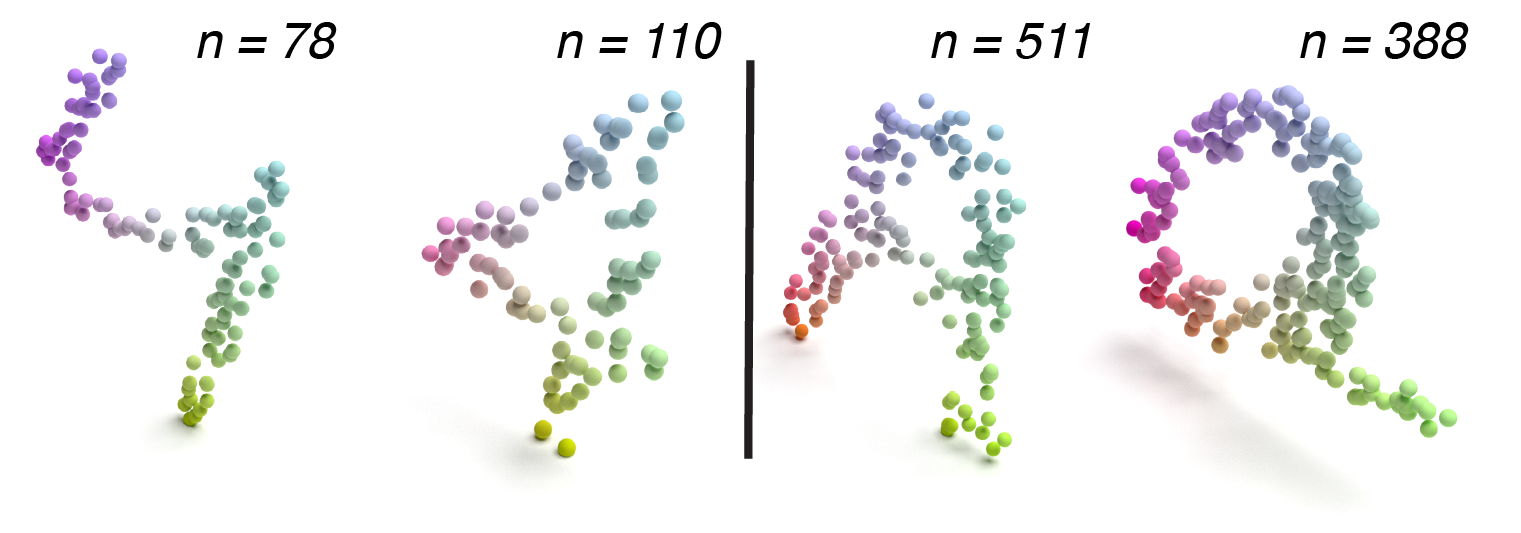}
  \vspace{-4mm}
  \caption{Generated distributions from MNIST and Letters datasets using WFM. Each distribution is realized as a point-cloud with naturally varying sample sizes ($n$), demonstrating WFM's unique ability to learn from and generate distributions without requiring fixed-size realizations.} \label{fig:wfm_mnist}
  \vspace{-3mm}
\end{figure}

Another common instance of BW manifolds arising in single-cell genomics is through aggregating cells into common states. These clusters can be summarized by their mean gene expression and its covariance. On a scRNA-seq atlas elucidating human immune response to COVID \citep{stephenson2021single}, we combine cells into MetaCells \citep{persad2023seacells}, and quantify the gene expression mean and covariance for each. Here too we apply BW-FM conditioned on cell-state, which encompass the heterogeneity of immune profiles appearing as response to COVID infection. Despite the plurality of labels, BW-FM can synthesize correct examples for each condition; see \cref{fig:bw_metacells}.

\subsection{Families of general distributions}\label{subsec:exp_wfm}
Recall that for general distributions, we use entropic optimal transport to estimate McCann interpolations and velocities~\citep{cuturi2013sinkhorn}. With high-resolution CAD datasets, we samplr 2048-point realizations from each shape to approximate OT maps during training. For spatial transcriptomics, MNIST and Letters datasets, we work directly with their naturally varying sample sizes, leveraging entropic OT's ability to handle unequal point-cloud sizes while maintaining accurate transport estimation. These align with our assumption that the underlying distributions are continuous, with point-clouds serving as discrete realizations.

\begin{table}[h] 
\vspace{-1mm} 
\scalebox{0.80}{ 
\begin{tabular}{lcccccc} 
\toprule 
& \multicolumn{2}{c}{Airplane} & \multicolumn{2}{c}{Chair} & \multicolumn{2}{c}{Car} \\ 
& CD $\downarrow$ & EMD $\downarrow$ & CD $\downarrow$ & EMD $\downarrow$ & CD $\downarrow$ & EMD $\downarrow$\\ 
\toprule 
PointFlow & 75.68 & 70.74 & 62.84 & 60.57 & 58.10 & 56.25 \\ 
SoftFlow & 76.05 & 65.80 & 59.21 & 60.05 & 64.77 & 60.09 \\ 
DPF-Net & 75.18 & 65.55 & 62.00 & 58.53 & 62.35 & \cellcolor[HTML]{D0E0E3}54.48 \\ 
Shape-GF & 80.00 & 76.17 & 68.96 & 65.48 & 63.20 & 56.53 \\ 
PVD & \cellcolor[HTML]{FFE6CC}73.82 & \cellcolor[HTML]{FFE6CC}64.81 & \cellcolor[HTML]{D9EAD3}56.26 & \cellcolor[HTML]{D9EAD3}53.32 & \cellcolor[HTML]{D0E0E3}54.55 & \cellcolor[HTML]{D9EAD3}53.83 \\ 
PSF & \cellcolor[HTML]{D0E0E3}71.11 & \cellcolor[HTML]{D9EAD3}61.09 & \cellcolor[HTML]{FFE6CC}58.92 & \cellcolor[HTML]{D0E0E3}54.45 & \cellcolor[HTML]{FFE6CC}57.19 &  \cellcolor[HTML]{FFE6CC}56.07 \\ 
WFM (ours) & \cellcolor[HTML]{D9EAD3}69.88 & \cellcolor[HTML]{D0E0E3}64.44 & \cellcolor[HTML]{D0E0E3}57.62 & \cellcolor[HTML]{FFE6CC}57.93 & \cellcolor[HTML]{D9EAD3}53.41 & 58.10 \\ 
\end{tabular}} 
\vspace{-3mm} 
\caption{Performance evaluation using 1-NN Accuracy (Earth Mover's Distance and Chamfer Distance) on distributions from ShapeNet, realized as uniform 3D point-clouds. Top three results for each metric are highlighted (\textcolor[HTML]{9DC183}{1st}, \textcolor[HTML]{88C1D7}{2nd}, \textcolor[HTML]{FFBB7D}{3rd}). WFM achieves state-of-the-art CD performance on 2 datasets while maintaining competitive results across other metrics (data from~\citet{wu2023fast}), while producing diverse samples, see Figure \ref{fig:wfm_shapenet}.\looseness-1} 
\label{tab:3d_point_clouds} 
\vspace{-2mm} 
\end{table}

We compare WFM to many other generative models. Following in their footsteps, we measure generation quality based on the 1-Nearest-Neighbour accuracy metric between generated and test-set shapes. On uniform, 3D datasets, WFM is competitive with current approaches, but exemplifies itself with its unique ability to generate samples with varying sizes and in high-dimensions; see Tables \ref{tab:3d_point_clouds} \& \ref{tab:high_d_point_clouds}.

\subsubsection{2D \& 3D Shapes}\label{subsubsec:wfm_2D_3D}

Derived from 3D CAD designs, ShapeNet \& ModelNet~\citep{wu20153d, chang2015shapenet} are touchstone shape datasets in computational geometry.  Trained individually on samples from the \textit{chair}, \textit{car} and \textit{plane} classes of ShapeNet, WFM synthesized high quality shapes with diverse profiles and matches the performance of previous 3D generation algorithms; see \cref{fig:wfm_shapenet} and \cref{tab:3d_point_clouds}. Our framework's versatility allows for seamless integration of label information during training, enabling the synthesis of shapes conditioned on specific classes. On the full 40-class ModelNet dataset, WFM learned condition-dependent flows between distributions, enabling generation of diverse shapes from the same initial distribution based on the desired label; see \cref{fig:wfm_shapenet}. We stress that WFM is not restricted to only noisy source measures but can generate transformations between any two collections of distributions. To this end, we demonstrate that WFM can interpolate between two arbitrary elements in the dataset (e.g., between a lamp and a handbag) and complete shapes based on partial profiles (e.g., generate the remaining parts of a plane); see \cref{fig:wfm_mug2bowl}.

\begin{table}[h]
\centering
\scalebox{0.85}{
\begin{tabular}{l|cc|cc}
\multicolumn{1}{c}{} & \multicolumn{2}{c|}{Current methods} & \multicolumn{2}{c}{WFM (ours)} \\
& CD $\downarrow$ & EMD $\downarrow$ & CD $\downarrow$ & EMD $\downarrow$ \\
\hline
\multicolumn{5}{l}{\textbf{Variable-sized realizations}} \\
\hline
MNIST ($4$) & NA & NA & 63.34 & 59.97 \\
Letters ($A$) & NA & NA & 62.12 & 58.68 \\
\hline
\multicolumn{5}{l}{\textbf{High-dimensional distributions}} \\
\hline
MERFISH & NA & NA & 52.80 & 54.52 \\
XENIUM & NA & NA & 60.69 & 64.20 \\
\end{tabular}}
\vspace{-2mm}
\caption{
1-Nearest-Neighbour Accuracy demonstrating WFM's unique capabilities. Using transformers and entropic transport maps, WFM handles both variable-sized realizations and high-dimensional distributions, unlike previous approaches limited to fixed-size, low-dimensional samples.
}\label{tab:high_d_point_clouds}
\vspace{-3mm}
\end{table}

Another novel facet of WFM is its ability to perform generative modeling from inhomogeneous datasets, where the number of points per point-cloud realization varies between independent samples. This happens in the MNIST or Letters datasets, where data is generated by thresholding low-resolution grayscale images. WFM sets itself apart from other methods, which are restricted to uniform datasets, by leveraging the entropic OT map's ability to compute feasible transformations between empirical samples of different sizes. Our experiments in \cref{fig:wfm_mnist} demonstrate that WFM generates high-quality \& diverse samples, despite large variability in the number of particles, itself a novel contribution.

\subsubsection{Spatial Transcriptomics}\label{subsubsec:wfm_high_d}

In spatial transcriptomics, a cell's niche is represented by the distribution of gene expression in its neighboring cells, which we observe through its immediate nearest neighbors in high-dimensional space. This approach is complementary to the BW representation of a niche (recall \cref{subsubsec:bw_real}), and serves as a more high fidelity view suited for fine-grain interactions. Due to their high dimensionality, these cellular microenvironment distributions have remained beyond the reach of generative models that depend on voxel-based neural networks. Instead, WFM uses transformers, whose permutation equivariance and indifference to dimensionality make them natural architectures for modeling these high-dimensional distributions through their point-cloud realizations \citep{haviv2024covariance}.

\begin{figure}[h]
  \centering
  \vspace{-3mm}
  \includegraphics[width=0.47\textwidth]{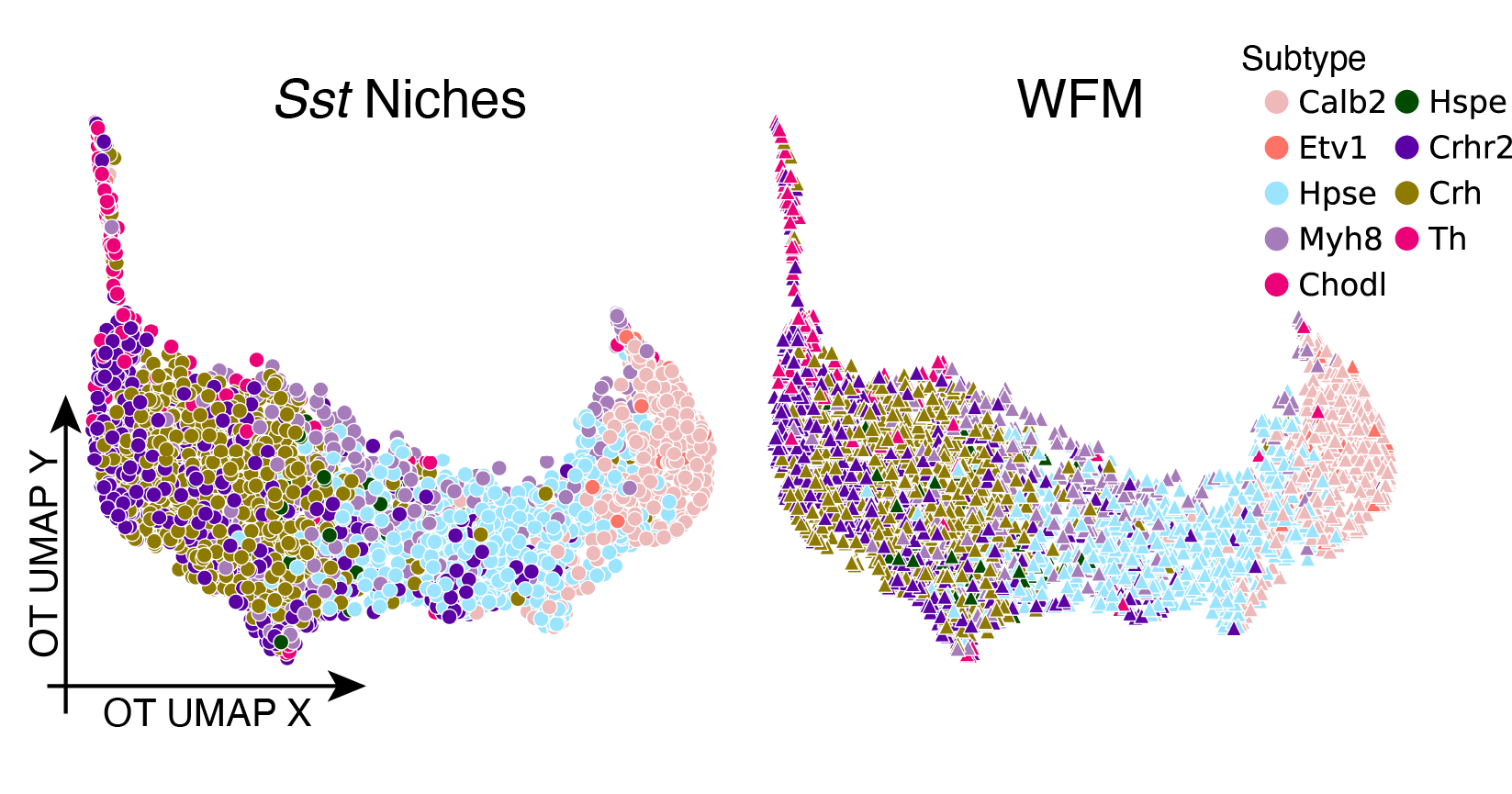}
  \vspace{-5mm}
    \caption{WFM enables high-dimensional generation of cellular microenvironments. (A) Wasserstein 2D embedding of observed \textit{Sst}-neuron niches in motor cortex, colored by subtype. (B) WFM generated niches faithfuly reproduce of the tissue landscape} \label{fig:wfm_sst_umap}
  \vspace{-3mm}
\end{figure}

\begin{figure}[h]
  \vspace{-1mm}
  \centering
  \includegraphics[width=0.44\textwidth]{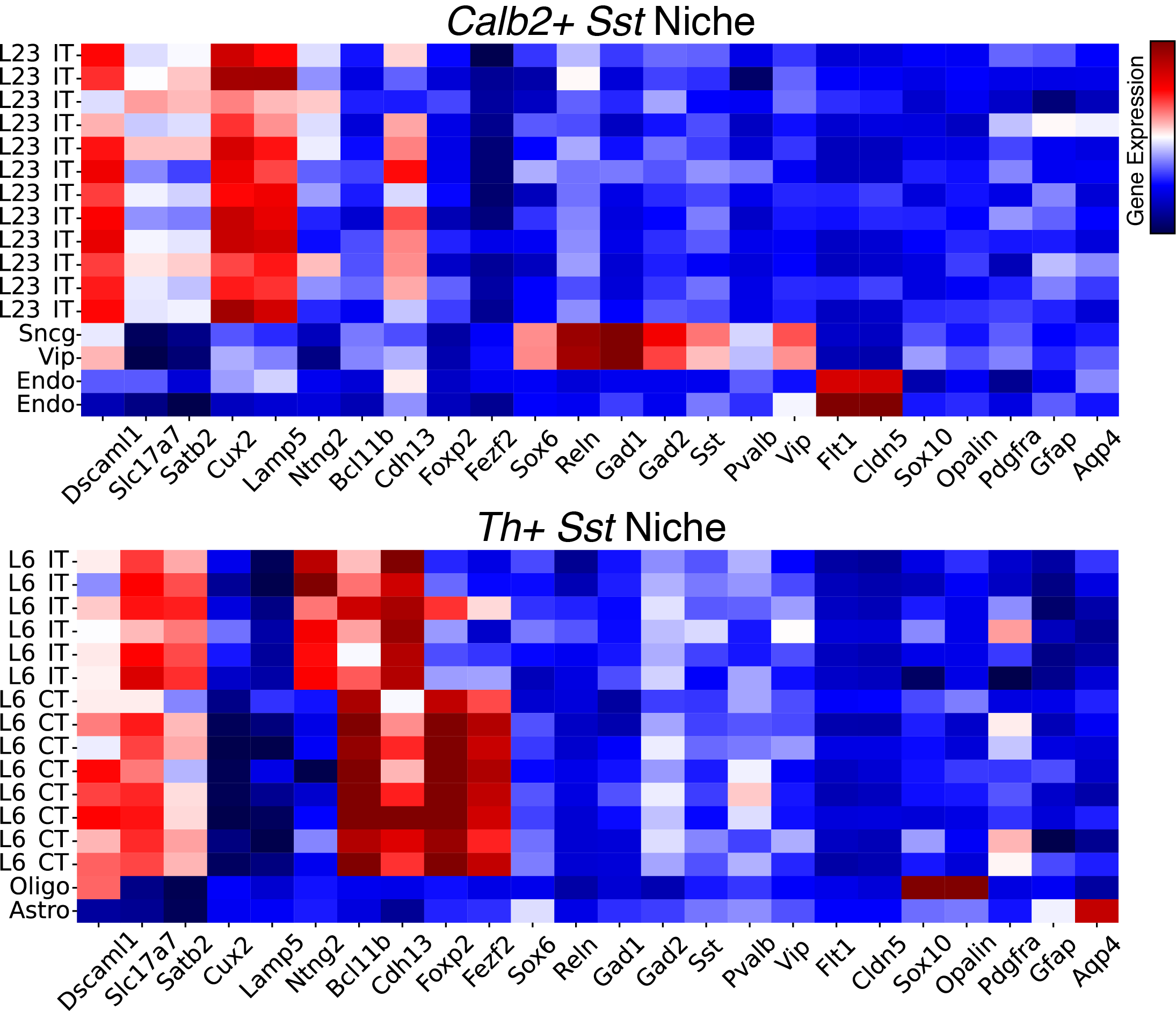}
  \vspace{-3mm}
    \caption{Examples of generated niches for \textit{Calb2}+ and  \textit{Th}+ \textit{Sst}-neurons (see \cref{fig:wfm_sst_umap}) display expected gene-expression signatures, with \textit{Calb2}+ niches enriched for upper-layer markers (\textit{Cux2}) and \textit{Th}+ for deep-layer markers (\textit{Ntng2} and \textit{Foxp2}). Each niche is represented as a high-dimensional distribution realized through point-clouds, demonstrating WFM's unique capability to generate distributions in spaces where previous methods, limited by voxelization, cannot operate.} \label{fig:wfm_sst_heatmaps}
  \vspace{-5mm}
\end{figure}

In the motor cortex, somatostatin (\textit{Sst}) interneurons are organized into phenotypically distinct subtypes, each localized to specific cortical layers \citep{wu2023cortical}. Using the 254-gene MERFISH atlas \citep{zhang2021spatially}, we represent each \textit{Sst} neuron's microenvironment as a distribution in gene expression space, observed through its 16 nearest neighbors. Despite operating in this high-dimensional space, WFM successfully generates niche distributions that capture both global tissue organization and subtype-specific molecular signatures; see Figures \ref{fig:wfm_sst_umap} \& \ref{fig:wfm_sst_heatmaps}.

\section{Conclusion and outlook}\label{sec:conclusion}
This work shows how to appropriately \textit{lift} the Riemannian flow matching paradigm of \citet{chen2023riemannian} to the Wasserstein space, resulting in Wasserstein flow matching. Our motivations stem from modern datasets, where each data sample is a probability distribution, necessitating this extension for generative modeling purposes. Our contributions are algorithmic in nature, which incorporate various elements, such as estimating optimal transport maps via entropic OT, closed-form expressions over the Bures--Wasserstein space, and attention mechanisms in neural network architectures. Our algorithm is capable of generating realistic data from Gaussian and general distribution realized as variable-size or high-dimensional point-clouds. Both contexts are highly relevant in single-cell transcriptomics for synthesizing of microenvironments 
and cellular states.

\section{Impact Statement}

Wasserstein Flow Matching is designed for generative modeling of distributions, with applications in computational biology. As with any generative model, care should be taken when using WFM to draw biological conclusions.

\section{Acknowledgements}

AAP acknowledges support from NSF grant DMS-1922658. DP is an Investigator at the Howard Hughes Medical Institute and is supported by National Cancer Institute grants P30 CA08748 and U54 CA209975.

\bibliography{wfm}

\begin{thebibliography}{84}
\providecommand{\natexlab}[1]{#1}
\providecommand{\url}[1]{\texttt{#1}}
\expandafter\ifx\csname urlstyle\endcsname\relax
  \providecommand{\doi}[1]{doi: #1}\else
  \providecommand{\doi}{doi: \begingroup \urlstyle{rm}\Url}\fi

\bibitem[Achlioptas et~al.(2018)Achlioptas, Diamanti, Mitliagkas, and
  Guibas]{achlioptas2018learning}
Achlioptas, P., Diamanti, O., Mitliagkas, I., and Guibas, L.
\newblock Learning representations and generative models for 3d point
  clouds--supplementary material--.
\newblock \emph{arXiv preprint arXiv:1801.07829}, 2018.

\bibitem[Albergo \& Vanden-Eijnden(2022)Albergo and
  Vanden-Eijnden]{albergo2022building}
Albergo, M.~S. and Vanden-Eijnden, E.
\newblock Building normalizing flows with stochastic interpolants.
\newblock \emph{arXiv preprint arXiv:2209.15571}, 2022.

\bibitem[Altschuler et~al.(2017)Altschuler, Weed, and Rigollet]{AltWeeRig17}
Altschuler, J., Weed, J., and Rigollet, P.
\newblock Near-linear time approximation algorithms for optimal transport via
  {S}inkhorn iteration.
\newblock In \emph{Advances in Neural Information Processing Systems 30: Annual
  Conference on Neural Information Processing Systems 2017, 4-9 December 2017,
  Long Beach, CA, {USA}}, 2017.

\bibitem[Altschuler et~al.(2021)Altschuler, Chewi, Gerber, and
  Stromme]{altschuler2021averaging}
Altschuler, J., Chewi, S., Gerber, P.~R., and Stromme, A.
\newblock Averaging on the {B}ures--{W}asserstein manifold: {D}imension-free
  convergence of gradient descent.
\newblock \emph{Advances in Neural Information Processing Systems},
  34:\penalty0 22132--22145, 2021.

\bibitem[Amari(2016)]{amari2016information}
Amari, S.-i.
\newblock \emph{Information geometry and its applications}, volume 194.
\newblock Springer, 2016.

\bibitem[Ambrosio et~al.(2008)Ambrosio, Gigli, and Savar\'{e}]{Ambrosio2008}
Ambrosio, L., Gigli, N., and Savar\'{e}, G.
\newblock \emph{Gradient flows in metric spaces and in the space of probability
  measures}.
\newblock Lectures in Mathematics ETH Z\"{u}rich. Birkh\"{a}user Verlag, Basel,
  second edition, 2008.
\newblock ISBN 978-3-7643-8721-1.

\bibitem[Amos et~al.(2022)Amos, Cohen, Luise, and Redko]{amos2022meta}
Amos, B., Cohen, S., Luise, G., and Redko, I.
\newblock Meta optimal transport.
\newblock \emph{arXiv preprint arXiv:2206.05262}, 2022.

\bibitem[Atanackovic et~al.(2024)Atanackovic, Zhang, Amos, Blanchette, Lee,
  Bengio, Tong, and Neklyudov]{atanackovic2024meta}
Atanackovic, L., Zhang, X., Amos, B., Blanchette, M., Lee, L.~J., Bengio, Y.,
  Tong, A., and Neklyudov, K.
\newblock Meta flow matching: Integrating vector fields on the wasserstein
  manifold.
\newblock \emph{arXiv preprint arXiv:2408.14608}, 2024.

\bibitem[Ba(2016)]{ba2016layer}
Ba, J.~L.
\newblock Layer normalization.
\newblock \emph{arXiv preprint arXiv:1607.06450}, 2016.

\bibitem[Benamou \& Brenier(2000)Benamou and Brenier]{benamou2000computational}
Benamou, J.-D. and Brenier, Y.
\newblock A computational fluid mechanics solution to the
  {M}onge--{K}antorovich mass transfer problem.
\newblock \emph{Numerische Mathematik}, 84\penalty0 (3):\penalty0 375--393,
  2000.

\bibitem[Bennett(2010)]{bennett2010openstreetmap}
Bennett, J.
\newblock \emph{OpenStreetMap}.
\newblock Packt Publishing Ltd, 2010.

\bibitem[Binnewies et~al.(2018)Binnewies, Roberts, Kersten, Chan, Fearon,
  Merad, Coussens, Gabrilovich, Ostrand-Rosenberg, Hedrick,
  et~al.]{binnewies2018understanding}
Binnewies, M., Roberts, E.~W., Kersten, K., Chan, V., Fearon, D.~F., Merad, M.,
  Coussens, L.~M., Gabrilovich, D.~I., Ostrand-Rosenberg, S., Hedrick, C.~C.,
  et~al.
\newblock Understanding the tumor immune microenvironment (time) for effective
  therapy.
\newblock \emph{Nature medicine}, 24\penalty0 (5):\penalty0 541--550, 2018.

\bibitem[Boeing(2017)]{boeing2017osmnx}
Boeing, G.
\newblock Osmnx: New methods for acquiring, constructing, analyzing, and
  visualizing complex street networks.
\newblock \emph{Computers, environment and urban systems}, 65:\penalty0
  126--139, 2017.

\bibitem[Bradbury et~al.(2021)Bradbury, Frostig, Hawkins, Johnson, Leary,
  Maclaurin, Necula, Paszke, VanderPlas, Wanderman-Milne,
  et~al.]{bradbury2021jax}
Bradbury, J., Frostig, R., Hawkins, P., Johnson, M.~J., Leary, C., Maclaurin,
  D., Necula, G., Paszke, A., VanderPlas, J., Wanderman-Milne, S., et~al.
\newblock Jax: Autograd and xla.
\newblock \emph{Astrophysics Source Code Library}, pp.\  ascl--2111, 2021.

\bibitem[Brenier(1991)]{Bre91}
Brenier, Y.
\newblock Polar factorization and monotone rearrangement of vector-valued
  functions.
\newblock \emph{Comm. Pure Appl. Math.}, 44\penalty0 (4):\penalty0 375--417,
  1991.

\bibitem[Bubeck et~al.(2023)Bubeck, Chandrasekaran, Eldan, Gehrke, Horvitz,
  Kamar, Lee, Lee, Li, Lundberg, et~al.]{bubeck2023sparks}
Bubeck, S., Chandrasekaran, V., Eldan, R., Gehrke, J., Horvitz, E., Kamar, E.,
  Lee, P., Lee, Y.~T., Li, Y., Lundberg, S., et~al.
\newblock Sparks of artificial general intelligence: {E}arly experiments with
  {GPT}-4.
\newblock \emph{arXiv preprint arXiv:2303.12712}, 2023.

\bibitem[Cai et~al.(2020)Cai, Yang, Averbuch-Elor, Hao, Belongie, Snavely, and
  Hariharan]{cai2020learning}
Cai, R., Yang, G., Averbuch-Elor, H., Hao, Z., Belongie, S., Snavely, N., and
  Hariharan, B.
\newblock Learning gradient fields for shape generation.
\newblock In \emph{Computer Vision--ECCV 2020: 16th European Conference,
  Glasgow, UK, August 23--28, 2020, Proceedings, Part III 16}, pp.\  364--381.
  Springer, 2020.

\bibitem[Carlen \& Gangbo(2003)Carlen and Gangbo]{carlen2003constrained}
Carlen, E.~A. and Gangbo, W.
\newblock Constrained steepest descent in the 2-{W}asserstein metric.
\newblock \emph{Annals of mathematics}, pp.\  807--846, 2003.

\bibitem[Chang et~al.(2015)Chang, Funkhouser, Guibas, Hanrahan, Huang, Li,
  Savarese, Savva, Song, Su, et~al.]{chang2015shapenet}
Chang, A.~X., Funkhouser, T., Guibas, L., Hanrahan, P., Huang, Q., Li, Z.,
  Savarese, S., Savva, M., Song, S., Su, H., et~al.
\newblock Shapenet: An information-rich 3d model repository.
\newblock \emph{arXiv preprint arXiv:1512.03012}, 2015.

\bibitem[Chen \& Lipman(2023)Chen and Lipman]{chen2023riemannian}
Chen, R.~T. and Lipman, Y.
\newblock Riemannian flow matching on general geometries.
\newblock \emph{arXiv preprint arXiv:2302.03660}, 2023.

\bibitem[Chen et~al.(2024)Chen, Goldstein, Hua, Albergo, Boffi, and
  Vanden-Eijnden]{chen2024probabilistic}
Chen, Y., Goldstein, M., Hua, M., Albergo, M.~S., Boffi, N.~M., and
  Vanden-Eijnden, E.
\newblock Probabilistic forecasting with stochastic interpolants and
  {F}{\"o}llmer processes.
\newblock \emph{arXiv preprint arXiv:2403.13724}, 2024.

\bibitem[Cheng et~al.(2024)Cheng, Li, Peng, and Liu]{cheng2024categorical}
Cheng, C., Li, J., Peng, J., and Liu, G.
\newblock Categorical flow matching on statistical manifolds.
\newblock \emph{arXiv preprint arXiv:2405.16441}, 2024.

\bibitem[Cui et~al.(2024)Cui, Wang, Maan, Pang, Luo, Duan, and
  Wang]{cui2024scgpt}
Cui, H., Wang, C., Maan, H., Pang, K., Luo, F., Duan, N., and Wang, B.
\newblock scgpt: toward building a foundation model for single-cell multi-omics
  using generative ai.
\newblock \emph{Nature Methods}, pp.\  1--11, 2024.

\bibitem[Cuturi(2013)]{cuturi2013sinkhorn}
Cuturi, M.
\newblock {S}inkhorn distances: Lightspeed computation of optimal transport.
\newblock \emph{Advances in Neural Information Processing Systems}, 26, 2013.

\bibitem[Cuturi et~al.(2022)Cuturi, Meng-Papaxanthos, Tian, Bunne, Davis, and
  Teboul]{cuturi2022optimal}
Cuturi, M., Meng-Papaxanthos, L., Tian, Y., Bunne, C., Davis, G., and Teboul,
  O.
\newblock Optimal transport tools ({OTT}): {A} {JAX} toolbox for all things
  {W}asserstein.
\newblock \emph{arXiv preprint arXiv:2201.12324}, 2022.

\bibitem[Davis et~al.(2024)Davis, Kessler, Petrache, Bose,
  et~al.]{davis2024fisher}
Davis, O., Kessler, S., Petrache, M., Bose, A.~J., et~al.
\newblock Fisher flow matching for generative modeling over discrete data.
\newblock \emph{arXiv preprint arXiv:2405.14664}, 2024.

\bibitem[Deb et~al.(2021)Deb, Ghosal, and Sen]{deb2021rates}
Deb, N., Ghosal, P., and Sen, B.
\newblock Rates of estimation of optimal transport maps using plug-in
  estimators via barycentric projections.
\newblock \emph{Advances in Neural Information Processing Systems},
  34:\penalty0 29736--29753, 2021.

\bibitem[Divol et~al.(2022)Divol, Niles-Weed, and Pooladian]{divol2022optimal}
Divol, V., Niles-Weed, J., and Pooladian, A.-A.
\newblock Optimal transport map estimation in general function spaces.
\newblock \emph{arXiv preprint arXiv:2212.03722}, 2022.

\bibitem[Emami \& Pass(2025)Emami and Pass]{emami2025optimal}
Emami, P. and Pass, B.
\newblock Optimal transport with optimal transport cost: the
  {M}onge--{K}antorovich problem on {W}asserstein spaces.
\newblock \emph{Calculus of Variations and Partial Differential Equations},
  64\penalty0 (2):\penalty0 43, 2025.

\bibitem[Eyring et~al.(2024)Eyring, Klein, Uscidda, Palla, Kilbertus, Akata,
  and Theis]{eyring2024unbalancednessneuralmongemaps}
Eyring, L., Klein, D., Uscidda, T., Palla, G., Kilbertus, N., Akata, Z., and
  Theis, F.
\newblock Unbalancedness in neural monge maps improves unpaired domain
  translation, 2024.
\newblock URL \url{https://arxiv.org/abs/2311.15100}.

\bibitem[Flamary et~al.(2021)Flamary, Courty, Gramfort, et~al.]{flamary2021pot}
Flamary, R., Courty, N., Gramfort, A., et~al.
\newblock Pot: Python optimal transport.
\newblock \emph{Journal of Machine Learning Research}, 22\penalty0
  (78):\penalty0 1--8, 2021.
\newblock URL \url{http://jmlr.org/papers/v22/20-451.html}.

\bibitem[Gadelha et~al.(2018)Gadelha, Wang, and
  Maji]{gadelha2018multiresolution}
Gadelha, M., Wang, R., and Maji, S.
\newblock Multiresolution tree networks for 3d point cloud processing.
\newblock In \emph{Proceedings of the European Conference on Computer Vision
  (ECCV)}, pp.\  103--118, 2018.

\bibitem[Gayoso et~al.(2022)Gayoso, Lopez, Xing, Boyeau, Valiollah Pour~Amiri,
  Hong, Wu, Jayasuriya, Mehlman, Langevin, et~al.]{gayoso2022python}
Gayoso, A., Lopez, R., Xing, G., Boyeau, P., Valiollah Pour~Amiri, V., Hong,
  J., Wu, K., Jayasuriya, M., Mehlman, E., Langevin, M., et~al.
\newblock A python library for probabilistic analysis of single-cell omics
  data.
\newblock \emph{Nature biotechnology}, 40\penalty0 (2):\penalty0 163--166,
  2022.

\bibitem[Gelbrich(1990)]{gelbrich1990formula}
Gelbrich, M.
\newblock On a formula for the l2 {W}asserstein metric between measures on
  {E}uclidean and {H}ilbert spaces.
\newblock \emph{Mathematische Nachrichten}, 147\penalty0 (1):\penalty0
  185--203, 1990.

\bibitem[Haviv et~al.(2024{\natexlab{a}})Haviv, Kunes, Dougherty, Burdziak,
  Nawy, Gilbert, and Pe’Er]{haviv2024wasserstein}
Haviv, D., Kunes, R.~Z., Dougherty, T., Burdziak, C., Nawy, T., Gilbert, A.,
  and Pe’Er, D.
\newblock Wasserstein wormhole: Scalable optimal transport distance with
  transformers.
\newblock \emph{ArXiv}, 2024{\natexlab{a}}.

\bibitem[Haviv et~al.(2024{\natexlab{b}})Haviv, Rem{\v{s}}{\'\i}k, Gatie,
  Snopkowski, et~al.]{haviv2024covariance}
Haviv, D., Rem{\v{s}}{\'\i}k, J., Gatie, M., Snopkowski, C., et~al.
\newblock The covariance environment defines cellular niches for spatial
  inference.
\newblock \emph{Nature Biotechnology}, pp.\  1--12, 2024{\natexlab{b}}.

\bibitem[H{\"u}tter \& Rigollet(2021)H{\"u}tter and
  Rigollet]{hutter2021minimax}
H{\"u}tter, J.-C. and Rigollet, P.
\newblock Minimax estimation of smooth optimal transport maps.
\newblock \emph{The Annals of Statistics}, 49\penalty0 (2):\penalty0
  1166--1194, 2021.

\bibitem[Jumper et~al.(2021)Jumper, Evans, Pritzel, Green, Figurnov,
  Ronneberger, Tunyasuvunakool, Bates, {\v{Z}}{\'\i}dek, Potapenko,
  et~al.]{jumper2021highly}
Jumper, J., Evans, R., Pritzel, A., Green, T., Figurnov, M., Ronneberger, O.,
  Tunyasuvunakool, K., Bates, R., {\v{Z}}{\'\i}dek, A., Potapenko, A., et~al.
\newblock Highly accurate protein structure prediction with alphafold.
\newblock \emph{nature}, 596\penalty0 (7873):\penalty0 583--589, 2021.

\bibitem[Kim et~al.(2020)Kim, Lee, Kang, Lee, and Kim]{kim2020softflow}
Kim, H., Lee, H., Kang, W.~H., Lee, J.~Y., and Kim, N.~S.
\newblock Softflow: Probabilistic framework for normalizing flow on manifolds.
\newblock \emph{Advances in Neural Information Processing Systems},
  33:\penalty0 16388--16397, 2020.

\bibitem[Kingma(2014)]{kingma2014adam}
Kingma, D.~P.
\newblock Adam: A method for stochastic optimization.
\newblock \emph{arXiv preprint arXiv:1412.6980}, 2014.

\bibitem[Klein et~al.(2024)Klein, Uscidda, Theis, and
  Cuturi]{klein2024genotentropicgromovwasserstein}
Klein, D., Uscidda, T., Theis, F., and Cuturi, M.
\newblock Genot: Entropic (gromov) wasserstein flow matching with applications
  to single-cell genomics, 2024.
\newblock URL \url{https://arxiv.org/abs/2310.09254}.

\bibitem[Klokov et~al.(2020)Klokov, Boyer, and Verbeek]{klokov2020discrete}
Klokov, R., Boyer, E., and Verbeek, J.
\newblock Discrete point flow networks for efficient point cloud generation.
\newblock In \emph{European Conference on Computer Vision}, pp.\  694--710.
  Springer, 2020.

\bibitem[Lambert et~al.(2022)Lambert, Chewi, Bach, Bonnabel, and
  Rigollet]{lambert2022variational}
Lambert, M., Chewi, S., Bach, F., Bonnabel, S., and Rigollet, P.
\newblock Variational inference via {W}asserstein gradient flows.
\newblock \emph{Advances in Neural Information Processing Systems},
  35:\penalty0 14434--14447, 2022.

\bibitem[Lee et~al.(2019)Lee, Lee, Kim, Kosiorek, Choi, and Teh]{lee2019set}
Lee, J., Lee, Y., Kim, J., Kosiorek, A., Choi, S., and Teh, Y.~W.
\newblock Set transformer: A framework for attention-based
  permutation-invariant neural networks.
\newblock In \emph{International conference on machine learning}, pp.\
  3744--3753. PMLR, 2019.

\bibitem[Lipman et~al.(2022)Lipman, Chen, Ben-Hamu, Nickel, and
  Le]{lipman2022flow}
Lipman, Y., Chen, R.~T., Ben-Hamu, H., Nickel, M., and Le, M.
\newblock Flow matching for generative modeling.
\newblock \emph{arXiv preprint arXiv:2210.02747}, 2022.

\bibitem[Liu et~al.(2022)Liu, Gong, and Liu]{liu2022flow}
Liu, X., Gong, C., and Liu, Q.
\newblock Flow straight and fast: Learning to generate and transfer data with
  rectified flow.
\newblock \emph{arXiv preprint arXiv:2209.03003}, 2022.

\bibitem[Lohoff et~al.(2022)Lohoff, Ghazanfar, Missarova, Koulena, Pierson,
  Griffiths, Bardot, Eng, Tyser, Argelaguet, et~al.]{lohoff2022integration}
Lohoff, T., Ghazanfar, S., Missarova, A., Koulena, N., Pierson, N., Griffiths,
  J.~A., Bardot, E.~S., Eng, C.-H., Tyser, R.~C., Argelaguet, R., et~al.
\newblock Integration of spatial and single-cell transcriptomic data elucidates
  mouse organogenesis.
\newblock \emph{Nature biotechnology}, 40\penalty0 (1):\penalty0 74--85, 2022.

\bibitem[Lopez et~al.(2018)Lopez, Regier, Cole, Jordan, and
  Yosef]{lopez2018deep}
Lopez, R., Regier, J., Cole, M.~B., Jordan, M.~I., and Yosef, N.
\newblock Deep generative modeling for single-cell transcriptomics.
\newblock \emph{Nature methods}, 15\penalty0 (12):\penalty0 1053--1058, 2018.

\bibitem[Manole et~al.(2021)Manole, Balakrishnan, Niles-Weed, and
  Wasserman]{manole2021plugin}
Manole, T., Balakrishnan, S., Niles-Weed, J., and Wasserman, L.
\newblock Plugin estimation of smooth optimal transport maps.
\newblock \emph{arXiv preprint arXiv:2107.12364}, 2021.

\bibitem[Mena \& Niles-Weed(2019)Mena and Niles-Weed]{mena2019statistical}
Mena, G. and Niles-Weed, J.
\newblock Statistical bounds for entropic optimal transport: sample complexity
  and the central limit theorem.
\newblock \emph{Advances in Neural Information Processing Systems}, 32, 2019.

\bibitem[Muzellec et~al.(2021)Muzellec, Vacher, Bach, Vialard, and
  Rudi]{muzellec2021near}
Muzellec, B., Vacher, A., Bach, F., Vialard, F.-X., and Rudi, A.
\newblock Near-optimal estimation of smooth transport maps with kernel
  sums-of-squares.
\newblock \emph{arXiv preprint arXiv:2112.01907}, 2021.

\bibitem[Nielsen(2020)]{nielsen2020elementary}
Nielsen, F.
\newblock An elementary introduction to information geometry.
\newblock \emph{Entropy}, 22\penalty0 (10):\penalty0 1100, 2020.

\bibitem[Nowotschin et~al.(2019)Nowotschin, Setty, Kuo, Liu, Garg, Sharma,
  Simon, Saiz, Gardner, Boutet, et~al.]{nowotschin2019emergent}
Nowotschin, S., Setty, M., Kuo, Y.-Y., Liu, V., Garg, V., Sharma, R., Simon,
  C.~S., Saiz, N., Gardner, R., Boutet, S.~C., et~al.
\newblock The emergent landscape of the mouse gut endoderm at single-cell
  resolution.
\newblock \emph{Nature}, 569\penalty0 (7756):\penalty0 361--367, 2019.

\bibitem[OpenAI(2022)]{dalle}
OpenAI.
\newblock Dall.e-2.
\newblock \url{https://openai.com/dall-e-2/}, 2022.

\bibitem[Otto(2001)]{otto2001geometry}
Otto, F.
\newblock The geometry of dissipative evolution equations: {T}he porous medium
  equation.
\newblock 2001.

\bibitem[Panaretos \& Zemel(2020)Panaretos and Zemel]{Panaretos2020}
Panaretos, V.~M. and Zemel, Y.
\newblock \emph{The Wasserstein Space}, pp.\  37--57.
\newblock Springer International Publishing, Cham, 2020.
\newblock ISBN 978-3-030-38438-8.

\bibitem[Persad et~al.(2023)Persad, Choo, Dien, Sohail, Masilionis,
  et~al.]{persad2023seacells}
Persad, S., Choo, Z.-N., Dien, C., Sohail, N., Masilionis, I., et~al.
\newblock Seacells infers transcriptional and epigenomic cellular states from
  single-cell genomics data.
\newblock \emph{Nature Biotechnology}, 41\penalty0 (12):\penalty0 1746--1757,
  2023.

\bibitem[Peyr{\'e} \& Cuturi(2019)Peyr{\'e} and Cuturi]{PeyCut19}
Peyr{\'e}, G. and Cuturi, M.
\newblock Computational optimal transport.
\newblock \emph{Foundations and Trends{\textregistered} in Machine Learning},
  11\penalty0 (5-6):\penalty0 355--607, 2019.

\bibitem[Pooladian \& Niles-Weed(2021)Pooladian and
  Niles-Weed]{pooladian2021entropic}
Pooladian, A.-A. and Niles-Weed, J.
\newblock Entropic estimation of optimal transport maps.
\newblock \emph{arXiv preprint arXiv:2109.12004}, 2021.

\bibitem[Pooladian et~al.(2022)Pooladian, Cuturi, and
  Niles-Weed]{pooladian2022debiaser}
Pooladian, A.-A., Cuturi, M., and Niles-Weed, J.
\newblock Debiaser beware: Pitfalls of centering regularized transport maps.
\newblock \emph{arXiv preprint arXiv:2202.08919}, 2022.

\bibitem[Pooladian et~al.(2023{\natexlab{a}})Pooladian, Ben-Hamu,
  Domingo-Enrich, Amos, Lipman, and Chen]{pooladian2023multisample}
Pooladian, A.-A., Ben-Hamu, H., Domingo-Enrich, C., Amos, B., Lipman, Y., and
  Chen, R.~T.
\newblock Multisample flow matching: {S}traightening flows with minibatch
  couplings.
\newblock \emph{arXiv preprint arXiv:2304.14772}, 2023{\natexlab{a}}.

\bibitem[Pooladian et~al.(2023{\natexlab{b}})Pooladian, Divol, and
  Niles-Weed]{pooladian2023minimax}
Pooladian, A.-A., Divol, V., and Niles-Weed, J.
\newblock Minimax estimation of discontinuous optimal transport maps: The
  semi-discrete case.
\newblock \emph{arXiv preprint arXiv:2301.11302}, 2023{\natexlab{b}}.

\bibitem[Rigollet \& Stromme(2022)Rigollet and Stromme]{rigollet2022sample}
Rigollet, P. and Stromme, A.~J.
\newblock On the sample complexity of entropic optimal transport.
\newblock \emph{arXiv preprint arXiv:2206.13472}, 2022.

\bibitem[Santambrogio(2015)]{San15}
Santambrogio, F.
\newblock Optimal transport for applied mathematicians.
\newblock \emph{Birk{\"a}user, NY}, 55\penalty0 (58-63):\penalty0 94, 2015.

\bibitem[Scetbon et~al.(2021)Scetbon, Cuturi, and Peyr{\'e}]{scetbon2021low}
Scetbon, M., Cuturi, M., and Peyr{\'e}, G.
\newblock Low-rank sinkhorn factorization.
\newblock In \emph{International Conference on Machine Learning}, pp.\
  9344--9354. PMLR, 2021.

\bibitem[Sinkhorn(1964)]{sinkhorn1964relationship}
Sinkhorn, R.
\newblock A relationship between arbitrary positive matrices and doubly
  stochastic matrices.
\newblock \emph{The annals of mathematical statistics}, 35\penalty0
  (2):\penalty0 876--879, 1964.

\bibitem[Song et~al.(2020)Song, Sohl-Dickstein, Kingma, Kumar, Ermon, and
  Poole]{song2020score}
Song, Y., Sohl-Dickstein, J., Kingma, D.~P., Kumar, A., Ermon, S., and Poole,
  B.
\newblock Score-based generative modeling through stochastic differential
  equations.
\newblock \emph{arXiv preprint arXiv:2011.13456}, 2020.

\bibitem[Stark et~al.(2024)Stark, Jing, Wang, Corso, Berger, Barzilay, and
  Jaakkola]{stark2024dirichlet}
Stark, H., Jing, B., Wang, C., Corso, G., Berger, B., Barzilay, R., and
  Jaakkola, T.
\newblock Dirichlet flow matching with applications to {DNA} sequence design.
\newblock \emph{arXiv preprint arXiv:2402.05841}, 2024.

\bibitem[Stephenson et~al.(2021)Stephenson, Reynolds, Botting, Calero-Nieto,
  Morgan, Tuong, Bach, Sungnak, Worlock, Yoshida, et~al.]{stephenson2021single}
Stephenson, E., Reynolds, G., Botting, R.~A., Calero-Nieto, F.~J., Morgan,
  M.~D., Tuong, Z.~K., Bach, K., Sungnak, W., Worlock, K.~B., Yoshida, M.,
  et~al.
\newblock Single-cell multi-omics analysis of the immune response in covid-19.
\newblock \emph{Nature medicine}, 27\penalty0 (5):\penalty0 904--916, 2021.

\bibitem[Theodoris et~al.(2023)Theodoris, Xiao, Chopra, Chaffin, Al~Sayed,
  Hill, Mantineo, Brydon, Zeng, Liu, et~al.]{theodoris2023transfer}
Theodoris, C.~V., Xiao, L., Chopra, A., Chaffin, M.~D., Al~Sayed, Z.~R., Hill,
  M.~C., Mantineo, H., Brydon, E.~M., Zeng, Z., Liu, X.~S., et~al.
\newblock Transfer learning enables predictions in network biology.
\newblock \emph{Nature}, 618\penalty0 (7965):\penalty0 616--624, 2023.

\bibitem[Tong et~al.(2023)Tong, Malkin, Huguet, Zhang, Rector-Brooks, Fatras,
  Wolf, and Bengio]{tong2023improving}
Tong, A., Malkin, N., Huguet, G., Zhang, Y., Rector-Brooks, J., Fatras, K.,
  Wolf, G., and Bengio, Y.
\newblock Improving and generalizing flow-based generative models with
  minibatch optimal transport.
\newblock \emph{arXiv preprint arXiv:2302.00482}, 2023.

\bibitem[Vaswani(2017)]{vaswani2017attention}
Vaswani, A.
\newblock Attention is all you need.
\newblock \emph{Advances in Neural Information Processing Systems}, 2017.

\bibitem[Villani(2009)]{Vil08}
Villani, C.
\newblock \emph{Optimal transport: old and new}, volume 338.
\newblock Springer, 2009.

\bibitem[Vyas et~al.(2023)Vyas, Shi, Le, Tjandra, Wu, Guo, Zhang, Zhang,
  Adkins, Ngan, et~al.]{vyas2023audiobox}
Vyas, A., Shi, B., Le, M., Tjandra, A., Wu, Y.-C., Guo, B., Zhang, J., Zhang,
  X., Adkins, R., Ngan, W., et~al.
\newblock Audiobox: {U}nified audio generation with natural language prompts.
\newblock \emph{arXiv preprint arXiv:2312.15821}, 2023.

\bibitem[Wang et~al.(2010)Wang, Ozolek, Slep{\v{c}}ev, Lee, Chen, and
  Rohde]{wang2010optimal}
Wang, W., Ozolek, J.~A., Slep{\v{c}}ev, D., Lee, A.~B., Chen, C., and Rohde,
  G.~K.
\newblock An optimal transportation approach for nuclear structure-based
  pathology.
\newblock \emph{IEEE transactions on medical imaging}, 30\penalty0
  (3):\penalty0 621--631, 2010.

\bibitem[Wu et~al.(2023{\natexlab{a}})Wu, Wang, Gong, Liu, Xiong, Ranjan,
  Krishnamoorthi, Chandra, and Liu]{wu2023fast}
Wu, L., Wang, D., Gong, C., Liu, X., Xiong, Y., Ranjan, R., Krishnamoorthi, R.,
  Chandra, V., and Liu, Q.
\newblock Fast point cloud generation with straight flows.
\newblock In \emph{Proceedings of the IEEE/CVF conference on computer vision
  and pattern recognition}, pp.\  9445--9454, 2023{\natexlab{a}}.

\bibitem[Wu et~al.(2023{\natexlab{b}})Wu, Sevier, Dwivedi, Saldi, Hairston, Yu,
  Abbott, Choi, Sherer, Qiu, et~al.]{wu2023cortical}
Wu, S.~J., Sevier, E., Dwivedi, D., Saldi, G.-A., Hairston, A., Yu, S., Abbott,
  L., Choi, D.~H., Sherer, M., Qiu, Y., et~al.
\newblock Cortical somatostatin interneuron subtypes form cell-type-specific
  circuits.
\newblock \emph{Neuron}, 111\penalty0 (17):\penalty0 2675--2692,
  2023{\natexlab{b}}.

\bibitem[Wu et~al.(2015)Wu, Song, Khosla, Yu, Zhang, Tang, and Xiao]{wu20153d}
Wu, Z., Song, S., Khosla, A., Yu, F., Zhang, L., Tang, X., and Xiao, J.
\newblock 3d shapenets: A deep representation for volumetric shapes.
\newblock In \emph{Proceedings of the IEEE conference on computer vision and
  pattern recognition}, pp.\  1912--1920, 2015.

\bibitem[Xing et~al.(2023)Xing, Feng, Chen, Dai, Hu, Xu, Wu, and
  Jiang]{xing2023survey}
Xing, Z., Feng, Q., Chen, H., Dai, Q., Hu, H., Xu, H., Wu, Z., and Jiang, Y.-G.
\newblock A survey on video diffusion models.
\newblock \emph{ACM Computing Surveys}, 2023.

\bibitem[Yang et~al.(2019)Yang, Huang, Hao, Liu, Belongie, and
  Hariharan]{yang2019pointflow}
Yang, G., Huang, X., Hao, Z., Liu, M.-Y., Belongie, S., and Hariharan, B.
\newblock Pointflow: 3d point cloud generation with continuous normalizing
  flows.
\newblock In \emph{Proceedings of the IEEE/CVF international conference on
  computer vision}, pp.\  4541--4550, 2019.

\bibitem[Zemel \& Panaretos(2019)Zemel and Panaretos]{zemel2019frechet}
Zemel, Y. and Panaretos, V.~M.
\newblock Fr{\'e}chet means and {P}rocrustes analysis in {W}asserstein space.
\newblock 2019.

\bibitem[Zhang et~al.(2022)Zhang, Tozzo, Higgins, and Ranganath]{zhang2022set}
Zhang, L., Tozzo, V., Higgins, J., and Ranganath, R.
\newblock Set norm and equivariant skip connections: Putting the deep in deep
  sets.
\newblock In \emph{International Conference on Machine Learning}, pp.\
  26559--26574. PMLR, 2022.

\bibitem[Zhang et~al.(2021)Zhang, Eichhorn, Zingg, Yao, Cotter, Zeng, Dong, and
  Zhuang]{zhang2021spatially}
Zhang, M., Eichhorn, S.~W., Zingg, B., Yao, Z., Cotter, K., Zeng, H., Dong, H.,
  and Zhuang, X.
\newblock Spatially resolved cell atlas of the mouse primary motor cortex by
  merfish.
\newblock \emph{Nature}, 598\penalty0 (7879):\penalty0 137--143, 2021.

\bibitem[Zhou et~al.(2021)Zhou, Du, and Wu]{zhou20213d}
Zhou, L., Du, Y., and Wu, J.
\newblock 3d shape generation and completion through point-voxel diffusion.
\newblock In \emph{Proceedings of the IEEE/CVF international conference on
  computer vision}, pp.\  5826--5835, 2021.

\end{thebibliography}
\bibliographystyle{icml2025}

\clearpage

\setcounter{figure}{0}

\makeatletter
\renewcommand{\thefigure}{S\@arabic\c@figure}
\makeatother

\newpage
\appendix
\onecolumn
\section{Entropic estimation of OT maps}\label{app:eot}

We briefly discuss how to estimate optimal transport maps between distributions using entropic OT. We refer the interested reader to \citet{pooladian2021entropic} for more information on this approach.

We first outline the numerical aspects of the approach; we follow \citet{PeyCut19}. Let $\mu = \sum_i m^{-1}\delta_{x_i}$ and $\nu = \sum_j n^{-1} \delta_{y_j}$ be two empirical distributions where $\bm{X} = \{x_1,\ldots,x_m\}$, $\bm{Y} = \{y_1,\ldots,y_n\}$. We first define the following polyhedral constraint set
\begin{align*}
    U_{m,n} \defeq \bigl\{P \in \R^{m\times n}_+ \ : \ P\bm{1}_m = m^{-1}\bm{1}_m\,, P^\top \bm{1}_n = n^{-1}\bm{1}_n\bigr\}\,,
\end{align*}
which represents the possible couplings between the two discrete measures. The entropic optimal transport coupling between the two discrete measures $\mu$ and $\nu$ is defined as the minimizer to the following strictly convex optimization problem
\begin{align}\label{eq:pi_star}
 \bm{P}^\star \defeq \argmin_{P \in U_{m,n}} \langle C, P\rangle + \eps H(P)\,,
\end{align}
where $\eps > 0$, $H(P) \defeq \sum_{i,j}P_{i,j}(\log(P_{i,j}) - 1)$, and $C_{i,j} \defeq \|x_i - y_j\|^2_2$. Sinkhorn's matrix scaling algorithm \citep{sinkhorn1964relationship} makes it possible to solve for $\bm P^\star$ with a runtime of $O(mn/\eps)$ \citep{AltWeeRig17}. We briefly stress three points:
\begin{enumerate}
    \item The coupling $\bm{P}^\star$ is \textit{not} a permutation matrix. The coupling lies inside the polytope $U_{m,n}$ and not at the vertices, and therefore is not a permutation matrix.
    \item When $\eps = 0$, the objective becomes a standard linear program with a runtime of $\tilde{O}(mn(m+n))$ (up to log factors) \citep[Chapter 3]{PeyCut19}. While we include a CPU implementation \citep{flamary2021pot} in the WFM codebase, this approach lacks GPU efficiency and substantially increases training time, making it impractical for most use cases.
    \item Instead, the regularization parameter $\eps$ serves as a tunable training hyperparameter. Lower $\eps$ values better approximate true the optimal transport map but require more Sinkhorn iterations for convergence, creating a direct trade-off between accuracy and computational efficiency.
\end{enumerate}
In all our experiments, we used the open-source package OTT-JAX (See \href{https://ott-jax.readthedocs.io/en/latest/}{https://ott-jax.readthedocs.io/en/latest/}) to compute the entropic coupling and the out-of-sample mapping \citep{cuturi2022optimal}.

\subsection{Rounded matchings}\label{subsec:rounded_matching}

Our first approach holds when $m=n$. In this case, we can greedily \textit{round} the noisy matching matrix $\bm P^\star$ to become a permutation. This is achieved through an iterative process of selecting the maximum value (argmax) and zeroing out corresponding rows and columns. This method repeatedly identifies the largest remaining probability, sets it to 1, and eliminates other entries in its row and column, ultimately resulting in a permutation matrix that preserves the probabilistic assignment implied by the original doubly stochastic matrix. This is merely a GPU-friendly heuristic approximation to the true optimal permutation matrix between the two point-clouds.

\subsection{Entropic transport map: An out-of-sample estimator}\label{subsec:out_of_sample}
A primal-dual relationship of the strictly convex program \eqref{eq:pi_star} shows that there exist vectors $(\bm{f}^\star,\bm{g}^\star) \in \R^{m}\times \R^n$ such that
\begin{align*}
    \bm P_{i,j}^\star = e^{\bm f_{i}^\star/\eps}e^{-\bm C_{i,j}/\eps}e^{\bm g_{j}^\star/\eps}
\end{align*}
These two vectors are called the Kantorovich potentials, which are initially defined on the support of $\mu$ and $\nu$, respectively. However, they can be readily extended to all of $\R^d$ \citep{mena2019statistical}, resulting in two functions
\begin{align*}
&\hat{f}(x) = -\eps \log \Bigl(\sum_{j=1}^n n^{-1} \exp((\bm g_j^\star - \|x-y_j\|^2 )/\eps)\Bigr)\,, \\
&\hat{g}(y) = -\eps \log \Bigl(\sum_{i=1}^m m^{-1} \exp((\bm f_i^\star - \|y-x_i\|^2 )/\eps)\Bigr)\,.
\end{align*}
Following \citet{pooladian2021entropic}, we can define the \textit{entropic transport map}, where the last equality is a simple calculation:
\begin{align}\label{eq:entmap}
    \hat{T}_\eps(x) &\defeq x - \nabla \hat{f}(x) = \frac{\sum_{j=1}^n y_j \exp((\bm g_j^\star - \|x-y_j\|^2 )/\eps)}{\sum_{j=1}^n \exp((\bm g_j^\star - \|x-y_j\|^2 )/\eps)}\,.
\end{align}
This estimator was initially to provide statistical approximations to the optimal transport map $T_\star^{\mu\to\nu}$ on the basis of samples; see \citet{pooladian2021entropic,rigollet2022sample,pooladian2023minimax,pooladian2022debiaser}.
Note that $\hat{T}_\eps(x)$ can be interpreted as the conditional expectation of the plan $\bm P^\star$ conditioned on out-of-sample inputs $x \in \R^d$, which is well-defined due to the relations above.  Finally, we stress that this estimator can be adapted to settings where the point-clouds $\mu$ and $\nu$ not only have different numbers of points, but also non-uniform weights. As this estimator is also a by-product of Sinkhorn's algorithm, it is also scalable and GPU-friendly.

\subsection{On the (statistical) approximations of geodesics}\label{sec:eot_stat}
We briefly collect a basic results pertaining to the (statistical) approximation of optimal transport paths. This  bound shows that the error grows along the trajectory, but is limited by the overall distance of the maps.
\begin{proposition}
Let $\mu,\nu$ be two probability measures and suppose $\mu$ has a density, and let $T^\star$ be the optimal transport map from $\mu$ to $\nu$. Let $\hat{T}$ be an estimator to the optimal transport map, defined with respect to data $X_1,\ldots,X_n\sim\mu$ and $Y_1,\ldots,Y_n \sim \nu$. Then for $t\in[0,1]$
\begin{align*}
    \E[W_2^2(\rho_t,\hat{\rho}_t)]\leq t^2 \E\|\hat{T}-T^\star\|^2_{L^2(\mu)}\,,
\end{align*}
where the outer expectation is taken with respect to the data, and we define
\begin{align*}
    \rho_t \defeq ((1-t){\rm{id}} + t T^\star)_\sharp\mu\,, \quad \hat{\rho}_t \defeq ((1-t){\rm{id}} + t \hat{T})_\sharp\mu\,
\end{align*}
\end{proposition}
\begin{proof}
The result follows immediately from a standard coupling argument to obtain  the linearized Wasserstein distance \citep{wang2010optimal,Panaretos2020}
\begin{align*}
    W_2^2(\rho_t,\hat{\rho}_t) &\leq \|((1-t){\rm{id}} + t T^\star) - ((1-t){\rm{id}} + t \hat{T})\|^2_{L^2(\mu)} = t^2\|\hat{T}-T^\star\|^2_{L^2(\mu)}\,.\qedhere
\end{align*}
\end{proof}

The entropic Brenier map is one particular estimator. We note two key properties of this map; see \citet{pooladian2021entropic} for in-depth discussions.
\begin{theorem}\label{thm:eot_main}
Suppose $\mu,\nu$ have density bounded above and below, and that the optimal transport map between them, denoted $T^\star$, is such that $(T^\star)^{-1}$ is at least twice differentiable and there exists $\lambda,\Lambda > 0$ such that
$$ \lambda I \preceq D T^\star \preceq \Lambda I\,. $$
Then, when estimated from $n$ samples from $\mu$ and $n$ samples from $\mu$, the entropic Brenier map has the following error
\begin{align}\label{eq:eot_main_decomp}
    \E\|\hat{T}_\eps - T^\star\|^2_{L^2(\mu)} \lesssim n^{-1/2}\log(n)\eps^{-d/2 - 1} + \eps^2\,,
\end{align}
where we suppress constants that depend on our assumptions. Performing a bias-variance trade-off in the regularization parameter, one obtains $\eps = \eps(n) \asymp n^{-1/(d+4)}$ and the total error becomes
\begin{align*}
    \E\|\hat{T}_\eps - T^\star\|^2_{L^2(\mu)} \lesssim_{\log(n)} n^{-2/(d+4)}\,.
\end{align*}
\end{theorem}

We emphasize that the assumptions in \cref{thm:eot_main} are standard in the literature \citep{hutter2021minimax,deb2021rates,muzellec2021near,divol2022optimal}. While the rate scales exponentially poorly with the dimension, we stress that existing lower bounds of estimation  (see \citet{hutter2021minimax}) also suffer from the curse of dimensionality, which is unavoidable for this task. Combining these two results, we can compare the geodesic given by the OT map, and the one induced by using the entropic map, where we write
\begin{align*}
    \hat{\rho}_t^\eps \defeq ((1-t)\rm{id} + t \hat{T}_\eps)_\sharp\mu\,.
\end{align*}
\begin{corollary}
Consider the same setting as \cref{thm:eot_main}. Then the geodesic given by the estimated entropic Brenier map, denoted by $\hat{\rho}_t^\eps$, on the basis of $n$ samples and $\eps \asymp n^{-1/(d+4)}$, is close to the true geodesic with the following error
\begin{align*}
    \E[W_2^2(\rho_t,\hat{\rho}_t^\eps)] \lesssim t^2 n^{-2/(d+4)}\,.
\end{align*}  
\end{corollary}

\section{Derivation of the WFM objective}\label{sec:wfm_derivation}

In this section, we give validity to the WFM for optimization purposes, and our choice of curves. For instance, recall the original Flow Matching objective \citep{lipman2022flow}
\begin{align}
    \cL_{\rm{FM}}(\theta) \defeq \int_0^1 \E_{X_t \sim p_t}\|f_\theta(X_t,t) - u_t(X_t)\|^2 \dd t\,,
\end{align}
where $f_\theta : \R^d \times [0,1] \to \R^d$ is a neural network, and $(p_t,u_t)_{t\in[0,1]}$ are a density-vector field pair that satisfy the continuity equation between two distribution $\mu$ and $\nu$.
\begin{align*}
    \partial p_t + \nabla\cdot(p_t u_t) = 0\,, \quad p_0 = \mu\,, p_1 = \nu\,.
\end{align*}
Note that, implicit in $\cL_{\rm{FM}}(\theta)$ is the endpoint constraints $\mu$ and $\nu$. Now, we average over possibly choices of $\mu \sim \mf p_0$ and $\nu \sim \mf p_1$, resulting in
\begin{align}
    \int_0^1 \iint \E_{X_t \sim p_t}\|f_\theta(X_t,t) - u_t(X_t)\|^2 \dd \mf p_0(\mu) \dd \mf p_1(\nu) \dd t\,.
\end{align}
As a particular case, take $(p_t,u_t) \gets (\mu_t,v_t)$, where the first argument is the McCann interpolation between $\mu$ and $\nu$, and $v_t$ is the optimal velocity field, which is a function of the optimal transport map from $\mu$ to $\nu$ (recall \cref{sec:wass_background}). This yields our final objective \eqref{eq:fm_wass}, which we recall here for convenience
\begin{align}
    \cL_{\rm{WFM}}(\theta) \defeq \int_0^1 \iint \E_{X_t \sim \mu_t}\|f_\theta(X_t,t) - v_t(X_t)\|^2 \dd \mf p_0(\mu) \dd \mf p_1(\nu) \dd t\,.
\end{align}

When $\mf p_0, \mf p_1$ are distributions over Gaussians, we have closed-form expressions for all objects of interest. When $\mf p_0, \mf p_1$ are distributions over general distributions, we approximate the geodesics based on point-cloud samples using entropic Brenier maps and their respective interpolations. We emphasize that the rounded-matching we employ (see \cref{subsec:rounded_matching}) is also a valid curve.

\subsection{Validity of Conditional Flows}\label{sec:cond_flow}
The key idea behind Flow Matching is that the (tractable) conditional velocity probability paths come together to form correct marginal velocities and paths from source distribution to target. In Riemannian Flow Matching, given a source and target $\mf p_1, \mf p_0\in \mathcal{P}(\mathcal{M})$ distributions over \textit{Riemannian} manifold $\mathcal{M}$, the conditional velocity field from $\mf p_0$ to a sample $\nu\sim\mf p_1$ is generated by flowing from each point $\mu\sim\mf p_0$ towards $\nu$ using the geodesic path between them. This produces the conditional probability path $\mf p_{t}(\cdot|\nu)$ generated by the vector field $\mf u_{t}(\cdot|\nu)$, which obey the continuity equation for \textit{Riemannian} manifolds:

\begin{align}
   {\partial_t}\mf p_{t}(\cdot|\nu) + \text{div}_{\mathcal{M}}(\mf p_{t}(\cdot|\nu)\mf u_t(\cdot|\nu)) = 0
\end{align}
where $\text{div}_{\mathcal{M}}$ is the Riemannian divergence operator. Following Theorem 1 in \citet{lipman2022flow} and equation 7 in \citet{chen2023riemannian}, the marginal vector field:
\begin{align}\label{eq:marginal_vf}
\mf u_t(\mu_t) = \int_{\mathcal{M}} \mf u_t(\mu_t|\nu)\frac{\mf p_t(\mu_t|\nu)}{\mf p_t(\mu_t)} \dd \mf p_{1}(\nu)
\end{align}
generates the marginal probability path:
\begin{align}\label{eq:marginal_prob}
\mf p_t(\mu_t) =\int_{\mathcal{M}} \mf p_t(\mu_t|\nu) \dd \mf p_{1}(\nu)
\end{align}
as $\mf p_t(\mu_t)$ and $\mf u_t(\mu_t)$ together obey the continuity equation. The proof requires regularity constraints on $\mf p_t(\mu_t|\nu)$ and $\mf u_t(\mu_t|\nu)$, which we assume as in \citet{chen2023riemannian}. A withstanding requirement that we need is that a unique optimal transport map exists between \textit{any} two measures $(\mu,\nu) \in \mf p_0 \times \mf p_1$. This can be ensured if, for example, all measures in $\mf p_0$ have a density, which is assumed throughout the text. Positivity of the path $\mu_t$ can be ensured through a myriad of conditions (e.g., source and target measures have positive density over the same support). In the explicit Gaussian-to-Gaussian case (i.e., the Bures--Wasserstein manifold), this assumption is trivially satisfied as the manifold is a \textit{genuine} finite-dimensional Riemannian manifold and the Riemannian machinery of \citet{chen2023riemannian} goes through with no modifications. 
We carry out the rest of our computations for general distributions over the Wasserstein space, and hence the computations are non-rigorous. For a fully rigorous treatment, see \citet{emami2025optimal}.

\subsubsection*{Conditional flows and paths for WFM}
To continue, we require the following proposition. Recall that a pre-metric is a function ${\rm{d}}:\cM \times \cM \to \R$ that satisfies the following three requirements
\begin{enumerate}
    \item ${\rm{d}}(x,y) \geq 0$ for all $x,y \in \cM$,
    \item ${\rm{d}}(x,y) = 0$ if and only if $x=y$,
    \item $\nabla_x {\rm{d}}(x,y) \neq 0$ if and only if $x\neq y$.
\end{enumerate}
\begin{lemma}\label{lem:wasserstein_metric}
The $2$-Wasserstein distance defines a valid pre-metric over $\cP_{2,\rm{ac}}(\R^d)$.
\end{lemma}
\begin{proof}
As the $2$-Wasserstein distance is a metric \citep{Vil08,San15}, the first and second conditions of a pre-metric are trivially satisfied. For the third condition, we note that for $\mu,\nu \in \cP_{2,\rm{ac}}(\R^d)$
\begin{align}\label{eq:wass_grad}
    \nabla \tfrac12 W_2^2(\cdot,\nu)(\mu) = -\log_{\mu}(\nu) = -2(T^{\mu\to\nu} - \rm{id})
\end{align}
For a proof of this computation, see \citet{altschuler2021averaging} or \citet{zemel2019frechet}. By chain-rule, we have that
\begin{align*}
    \nabla \tfrac12 W_2^2(\cdot,\nu)(\mu) = W_2(\mu,\nu)\nabla W_2(\cdot,\nu)(\mu) = -\log_{\mu}(\nu) 
\end{align*}
This latter quantity is only zero if and only if $\mu=\nu$, and thus the proof is complete.
\end{proof}

We now recall how to construct conditional vector fields as a function of pre-metrics ${\rm{d}}$, as described by Eq. (13) by \cite{chen2023riemannian}. The construction is the following
\begin{align}\label{eq:eq13_chen}
     {u}_t(x_t|y) &= (\tfrac{\dd}{\dd t}\ln(1-t)) {\rm{d}}(x_t,y) \nabla_x {\rm{d}}(x_t,y)/\|\nabla_x {\rm{d}}(x_t,y)\|^2_{g(x_t)}\,,
\end{align}
where $g$ is the metric on the tangent space at $x_t$, and $x_t$ is a geodesic in the form
\begin{align*}
    x_t = \exp_{x_0}((1-t)\log_{x_0}(x_1))\,.
\end{align*}
Note that our geodesics are precisely of the form
\begin{align*}
    \mu_t = \exp_\mu((1-t)\log_\mu(\nu))\,,
\end{align*}
where we recall $v\mapsto\exp_\mu(v) = ({\rm{id}}+v)_\sharp\mu$ and $\nu\mapsto\log_\mu(\nu) = T^{\mu\to\nu} - {\rm{id}}$. We now specialize these calculations to the case where the pre-metric is the  $2$-Wasserstein distance.
\begin{lemma}\label{lem:metric_to_vel}
Fix $\nu \sim \mathfrak p_1$. Let $\mu_t \mapsto \mathfrak u_t(\mu_t|\nu)$ denote the conditional vector field defined using \cref{eq:eq13_chen} with ${\rm{d}}$ replaced by the $2$-Wasserstein distance, evaluated at a geodesic $\mu_t$ between any $\mu \sim \mathfrak p_0$ and the fixed $\nu$. Then it holds that $\mathfrak u_t(\mu_t|\nu) = v_t$, where $v_t$ is given by \Cref{eq:vectorfield} between $\mu$ and $\nu$.
\end{lemma}
\begin{proof}
We compute, using \cref{eq:wass_grad},
\begin{align*}
    \mathfrak{u}_t(\mu_t|\nu) &= (\tfrac{\dd}{\dd t}\ln(1-t)) W_2(\mu_t,\nu) \nabla W_2(\mu_t,\nu)/\|\nabla W_2(\mu_t,\nu)\|^2_{L^2(\mu_t)} \\
    &= \frac{1}{1-t} (T^{\mu_t \to \nu} - {\rm{id}}) \\
    &= \frac{-1}{1-t}(T^{\mu\to\nu} \circ (T_t)^{-1} - {\rm{id}})\\
    &= \frac{1}{1-t}(T^{\mu\to\nu} - T_t )\circ (T_t)^{-1}\,, 
\end{align*}
where we make use of the fact that $T^{\mu_t\to\nu} = T^{\mu\to\nu}\circ (T^{\mu_t\to\nu})^{-1}$ (see e.g., Section 2 of \citet{carlen2003constrained}), and that $T_t = (1-t){\rm{id}} + t{T^{\mu\to\nu}}$. Simplifying, we obtain
\begin{align}\label{eq:mfut_new}
    \mathfrak{u}_t(\mu_t|\nu) = \frac{1}{1-t} (T^{\mu\to\nu} - (1-t){\rm{id}} - t T^{\mu\to\nu} )\circ (T_t)^{-1}  = (T^{\mu\to\nu} - {\rm{id}} )\circ (T_t)^{-1}\,,
\end{align}
which is precisely the optimal transport vector field $v_t$ stated in \eqref{eq:vectorfield}. 
\end{proof}
Together, we arrive at the following corollary, which validates the probability paths generated by our approach.
\begin{corollary}
The conditional flow $\mf p_t(\cdot|\nu)$ generated by the  conditional vector fields defined by \eqref{eq:mfut_new} satisfy $\mf p_1(\cdot|\nu) = \nu$. 
\end{corollary}
\begin{proof}
This follows immediately from Theorem 3.1 of \citet{chen2023riemannian} by passing through the pre-metric construction above. 
\end{proof}

\section{Multisample Wasserstein Flow Matching}\label{app:multisample}

Since optimal transport can be applied on the Wasserstein manifold itself, both WFM and BW-FM can be seamlessly integrated with the multisample FM (MS-FM) framework \citep{pooladian2023multisample,tong2023improving}. The core technique behind MS-FM is to use OT to match minibatches from source and target measures during training, rather than relying on random pairings. This has shown to improve learned flows while requiring fewer function evaluations to synthesize new samples. Applying MS-FM requires computing the pairwise distance matrix between source and target batch samples, denoted from $i \in \{1,\ldots,\texttt{Bsz}\}$. In the BW-FM setting, given two sets of Gaussians $\{(a_{i}, A_{i})\}_{i=1}^{\texttt{Bsz}}$ and $\{(b_{i}, B_{i})\}_{i=1}^{\texttt{Bsz}}$, their Frech{\'e}t ($W^{2}_{2}$) distance matrix is:

\begin{equation}\label{eq:frechet_dist}
 C_{i,j} = \|a_{i} - b_{j}\|^{2}_{2} + \text{Tr}(A_i+B_j - 2(A_i^{1/2}B_jA_i^{1/2})^{1/2})
\end{equation}

We then use entropic OT to approximately solve the assignment problem on $C$ and compute a transport matrix. This is the converted into a one-to-one assignment matrix via rounded matching (see \cref{subsec:rounded_matching}), ensuring the entire batch is used in training.

For general WFM, applying MS requires computing pairwise OT distance between all source and target samples within a minibatch. For large point-cloud samples, this is exorbitantly expensive, even with Sinkhorn iterations. For an efficient approximate, here too we rely on the Frech\'et distance, computed between empirical means and covariances of each point-cloud. Computation of the Frech\'et distance is markedly less resource-intensive than entropic OT, yet is notably correlated with EMD values (see Table 1 in \citet{haviv2024covariance}).

\section{Choice of source measure}\label{app:source_measure}

In both WFM and BW-FM, learning flows requires a source measure which is straightforward to sample from. For a source distribution on the space of $\{(m,\Sigma) : m \in \R^d, \Sigma \in \mathbb{S}^d_{++}\}$, we simply sample means and covariance matrices using independent Gaussian and Wishart distributions, respectively. By default, the parameters for the Gaussian component of the source matches the average and standard deviation of the means in the target, while the scale parameter in the Wishart is the barycenter of the data covariance.

To achieve high-quality generation of general distributions, it is essential that the initial (source) measure be diverse, rather than collapsed and degenerate. Indeed, while it is alluring to produce noise point-clouds by sampling particles from a single base distribution, i.e. $X = \{x_i\}_{i=1}^n, \, x_i \sim \mathcal{N}(0, I_d)$, as $n$ grows, the Wasserstein distance between empirical distributions goes to $0$.  To alleviate this, we draw point-clouds from multivariate Gaussians with a stochastic covariance:

\begin{align*}
L \sim \mathcal{N}(\mu_L,& \sigma_L \cdot I) \\
X = \{x_i\}_{i=1}^{n}, \,& x_i \sim \mathcal{N}(0, LL^T)
\end{align*}

where $\mu_L$ \& $\sigma_L$ are the average and standard deviation of the Cholesky factors from the empirical covariances of the target measure point-clouds. This ensures a wider source measure, producing a diverse range of noise point-clouds.

\begin{figure}[h]
  \centering
  \includegraphics[width=0.9\textwidth]{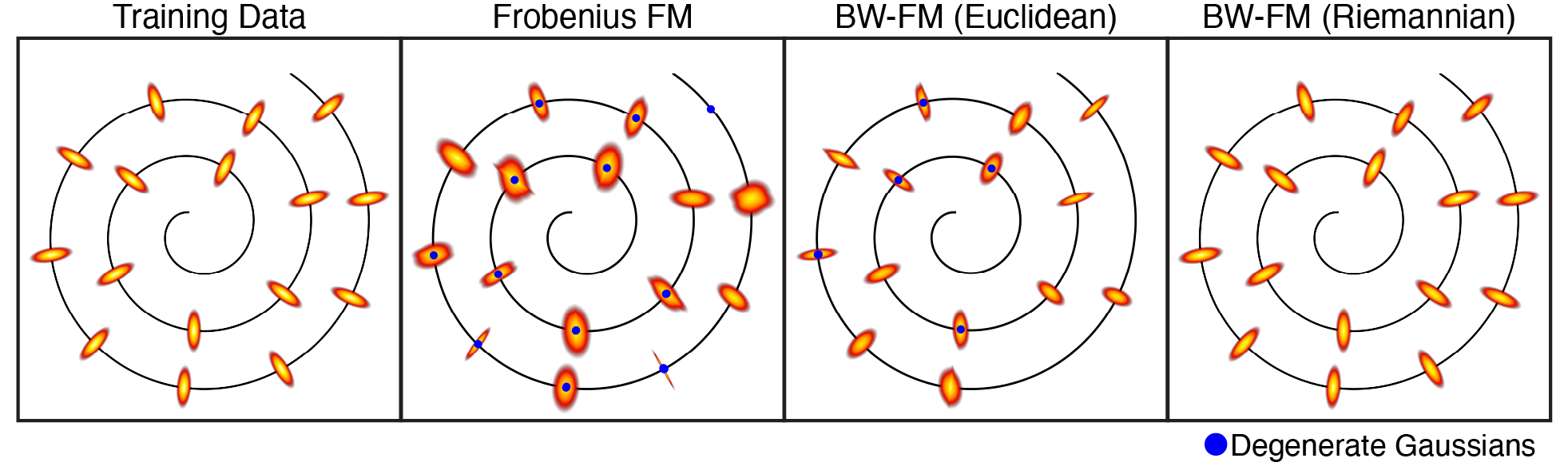}
  \vspace{-4mm}
  \caption{
  When the number of training examples is too few (in comparison to \cref{fig:bw_spirals}), all methods collapse on the training data, though only the Riemannian instantiation of BW-FM captures the covariances perfectly. 
  }\label{fig:bw_spirals_8}
\end{figure}

\section{Neural architecture \& training}\label{supp_sec:training}

\subsection{BW-FM on Gaussians}\label{app:BWFM_training}
The goal of BW-FM is to train a neural network to match the (Riemannian) time-derivative along the BW geodesics between Gaussians. The model employs a standard, fully connected neural network which takes as input concatenated values of $(m_{t},\Sigma_{t}, t)$ based on the McCann interpolation formula from \cref{sec:bw_background}. Since the covariance matrix is symmetric, only its lower-diagonal values are used, flattened into a vector of length $d(d+1)/2$. Time values are converted to Fourier features, an approach inspired by positional encodings in transformer literature \citep{vaswani2017attention}.  To streamline training, two separate networks are employed: one to match the time derivative of the mean $\dot{m}_{t}$ and another for the time derivative of the covariance matrix $\dot{\Sigma}^{{\rm{BW}}}_{t}$. The BW tangent norm is used as the loss function for training these networks.

By default, all models use a $6$-layer neural network using \textit{relu} non-linearity, with $1024$ neurons per layer, applying skip connections and layer-norm \cite{ba2016layer}. Training is performed for $100,000$ gradient descent steps using the Adam optimizer \citep{kingma2014adam} with an exponential learning rate decay of $0.97$ every $1000$ steps and batch size of $128$.

\subsection{WFM on general distributions}\label{app:wfm_training}
WFM estimates the optimal transport (OT) map for a given pair of interpolate point-cloud and time $(\bm{X}_{t}, t)$. Here too the time component $t$ is first converted into Fourier features. The model's architecture begins with an embedding layer, followed by a series of alternating multi-head attention and fully-connected layers. Skip connections and layer-norm are applied after each operation. The final layer projects the embeddings back to $X$'s original space using a dense layer with zero initialization. The model is trained by minimizing the squared distance between the predicted and true OT maps. \looseness-1

By default, the entropic OT map is constructed with regularization weight of $\varepsilon = 0.002$ and $200$ Sinkhorn iterations, which we found to be sufficient for convergence. Whenever the dataset consists of uniformly sized point-cloud realizations, we use rounded matching  (see \cref{subsec:rounded_matching}), otherwise we apply the out-of-sample estimator (see \cref{subsec:out_of_sample}) which can calculate maps between point-clouds with different sizes. The transformer network is composed to $6$ multi-head attention blocks, with an embedding dimension of $512$ and $4$ heads. Our model is optimizer with Adam \citep{kingma2014adam} using an exponential learning rate decay and batch size of $64$.

\begin{figure}[h]
  \centering
  \includegraphics[width=0.9\textwidth]{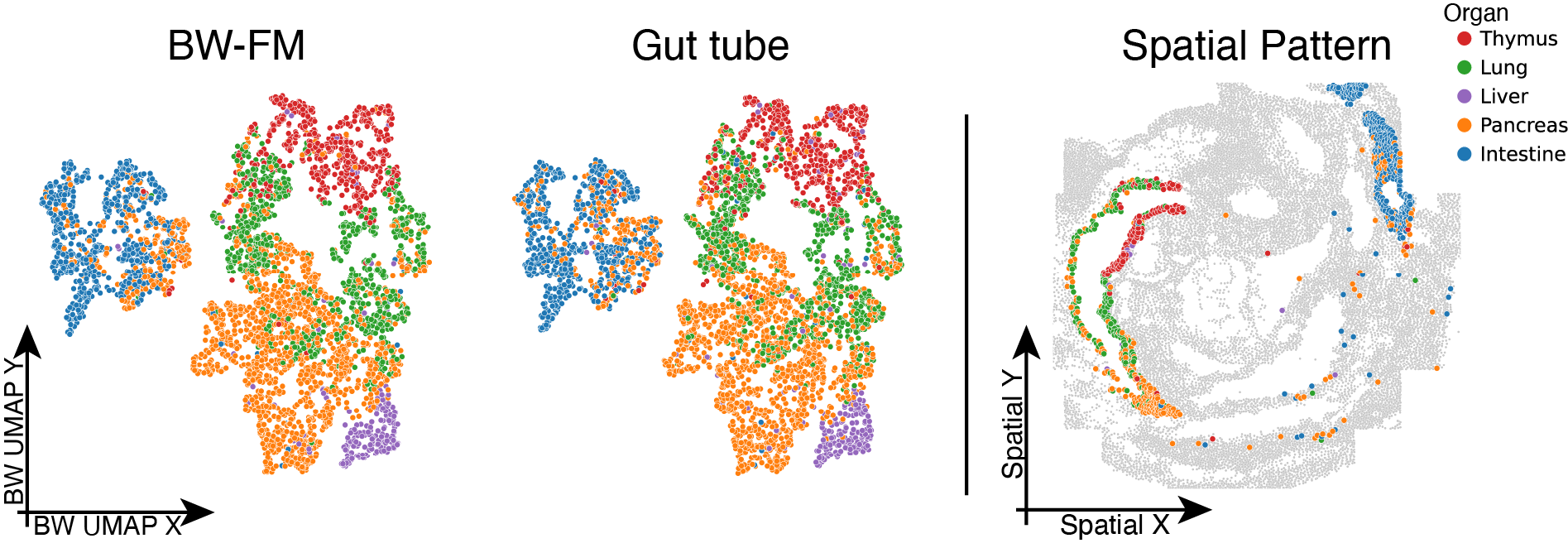}
  \vspace{-3mm}
  \caption{Comparing real and generated cellular microenvironments in embryonic gut tube development. Left: UMAP embedding of BW-FM generated (left panel) and observed (middle panel) microenvironments based on Bures-Wasserstein distance, colored by developing organ type. Right: Spatial arrangement of cells in the seqFISH embryo showing physical organization of organ development.}\label{fig:bw_gut_supp}
\end{figure}

WFM relies on JAX and OTT-JAX \citep{bradbury2021jax, cuturi2022optimal} and enjoys seamless optimization via end-to-end just-in-time compilation. For the ShapeNet experiments (see \cref{tab:3d_point_clouds}), the model is trained for 500,000 training steps, totaling to about 3 days of trainings of a single A100 GPU. All other experiments (see \cref{tab:high_d_point_clouds}) trained for 100,000 steps, requiring around 3-4 hours of GPU use. We note the Transformer's forward and backwards pass was the most significant source of computational overhead, as opposed to the Sinkhorn based approximation of OT maps. The computational complexity of both components is quadratic with point-cloud size, which limits the scalability of WFM. Making Transformers simpler to optimize and accelerating OT computations are both active areas within ML research \citep{amos2022meta, scetbon2021low, zhang2022set}, and future iterations of WFM can incorporate solution from those spaces.

\section{Experiment Details}

\subsection{Spatial Transcriptomics}

In our manuscript, we applied WFM and BW-FM on several spatial transcriptomics datasets, encompassing a variety of technologies and tissue contexts. From the 351-gene seqFISH embryogensis dataset, \citep{lohoff2022integration}, we focus on niches of gut-tube cells.  We compress gene-expression profiles down to their $16$ principal components (PC) and aggregate all the cells around each gut cell within an $80$ micron radius, yielding on average $29$ cells per niche. We then calculated the gene-expression PC mean and covariance within each environment to produce Gaussians for BW-FM. Generated Gaussians align with real data ad delineate budding organs into their appropriate regions (see \cref{fig:bw_gut_supp}).

\begin{figure}[h]
  \centering
  \includegraphics[width=0.85\textwidth]{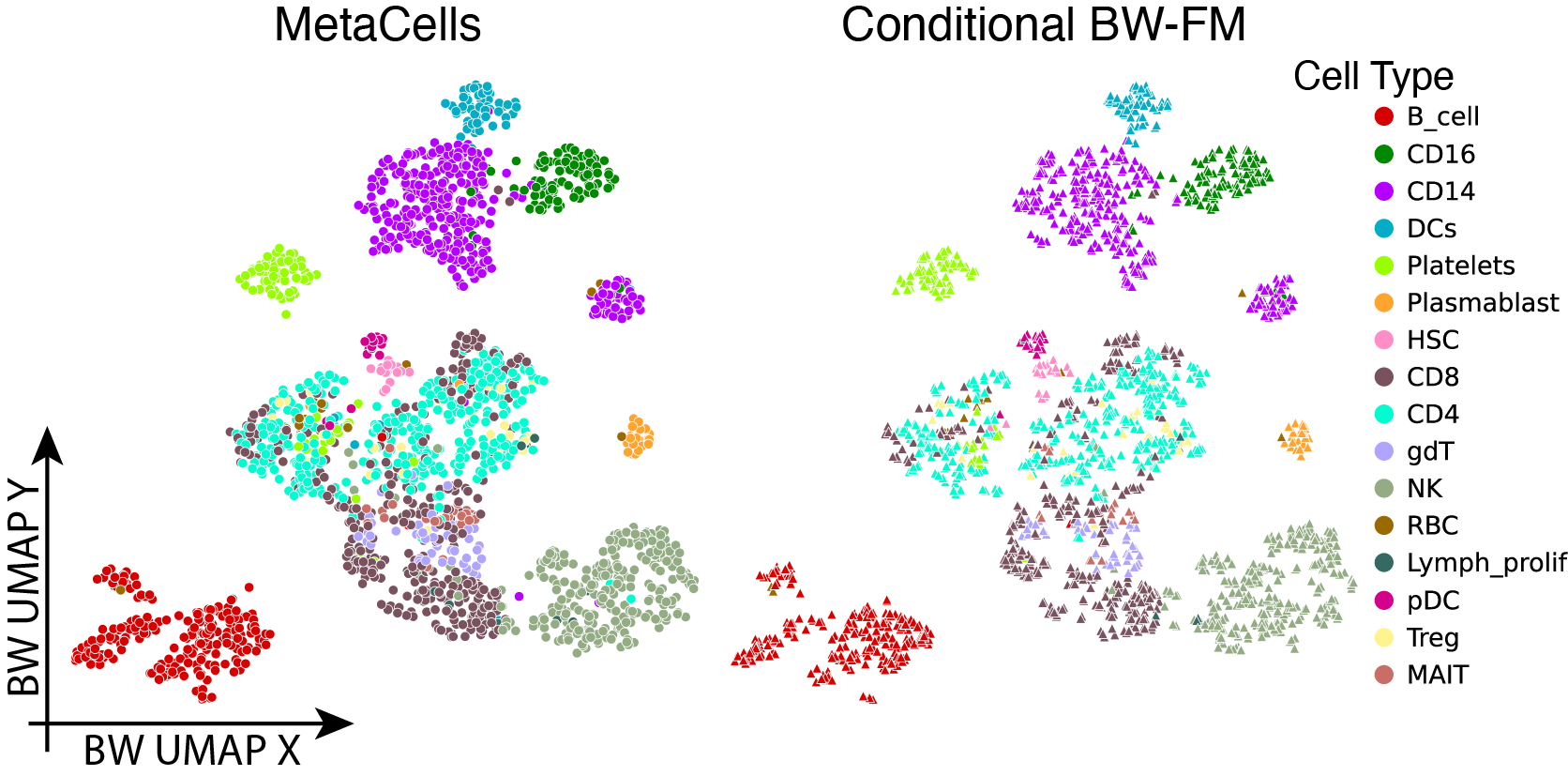}
  \caption{Conditional BW-FM applied to single-cell RNA sequencing data of immune response to COVID-19. Large scale single-cell atlases are commonly grouped into highly dedicated clusters called MetaCells \citep{persad2023seacells}. In this application, BW-FM is conditioned on cell state and trained to generate means and covariances of gene expression, focusing on the top 32 principal components, derived from aggregated cells. The model achieves high-quality sample generation, as evidenced by a label accuracy of 93.13\%.}\label{fig:bw_metacells}
\end{figure}

In a complementary approach, WFM is applied directly on gene-expression based point-clouds of niches, representing a realization of the continuous distribution of their environments. Uniquely suited for high-dimensional data, we apply WFM on a MERFISH atlas of the motor cortex XENIUM assay of melanoma metastasis to the brain \citep{haviv2024covariance}. In both dataset, we select the $k=16$ physical nearest neighbours of every cell, and aggregate expression profiles based on their first $16$ PCs.

We concentrated on \textit{Sst} interneurons in the motor cortex datas, which are divided into spatially segregated, phenotypically distinct subtypes. Applied unconditionally, WFM generated niches match the distribution of the real data based on EMD and CD 1-nearest-neighbour accuracy (see \cref{tab:high_d_point_clouds}).  We then assessed WFM's capability to comprehend  the relationship between cell state and environment and tasked it with conditional generation based on subtype label. Based on OT distances estimated via \textit{Wormhole} embeddings \citep{haviv2024wasserstein}, distributions generated by WFM recapitulated true organ environment. The label accuracy for WFM-generated data was 63.86\%, which was nearly identical to the test-set real data accuracy of 62.19\%.

\begin{figure}[h]
  \centering
  \includegraphics[width=0.7\textwidth]{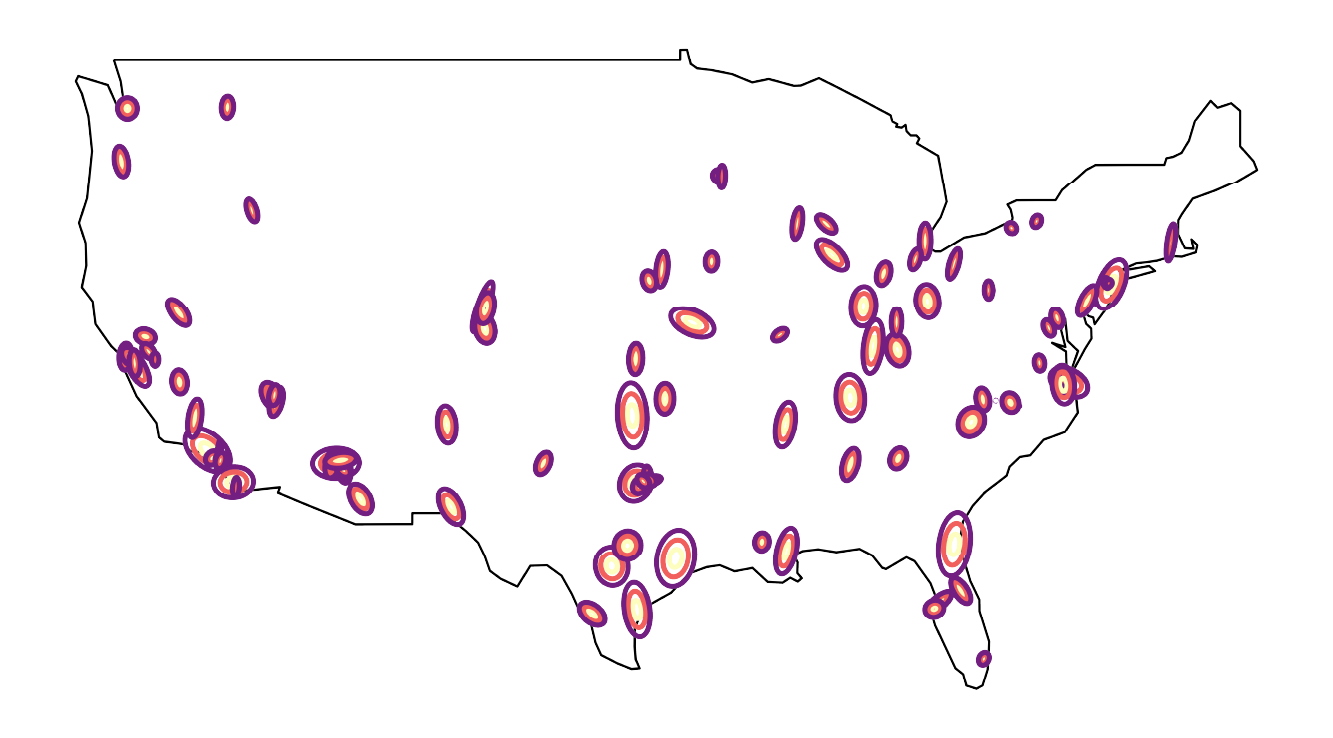}
  \vspace{-5mm}
  \caption{Cities Dataset. Gaussians representing the $100$ most populous cities in the continental US. The data was obtained from \cite{bennett2010openstreetmap} via OSMnx \citep{boeing2017osmnx}. The mean parameter is the longitude and latitude coordinate of each city and the covariance is the 2nd moment approximation of their metro area.}\label{fig:bw_cities}
  \vspace{-5mm}
\end{figure}

\section{2D \& 3D shapes}

The ShapeNet dataset consists of 3D shapes of $55$ different classes, Emulating the benchmarking effort in \citep{wu2023fast}, we apply WFM to generate $n=2048$ sized examples from the \textit{plane}, \textit{car} and \textit{chair} classes. At each gradient descent step, we sample $64$ shapes from the training set for each class, and randomly select $n=2048$ points from each. To evaluate generation quality, we synthesize point-clouds to much the size of the test set, and calculate the real or generated $1-NN$ accuracy based on EMD and CD metrics.

\begin{figure}[h]
  \centering
  \includegraphics[width=0.8\textwidth]{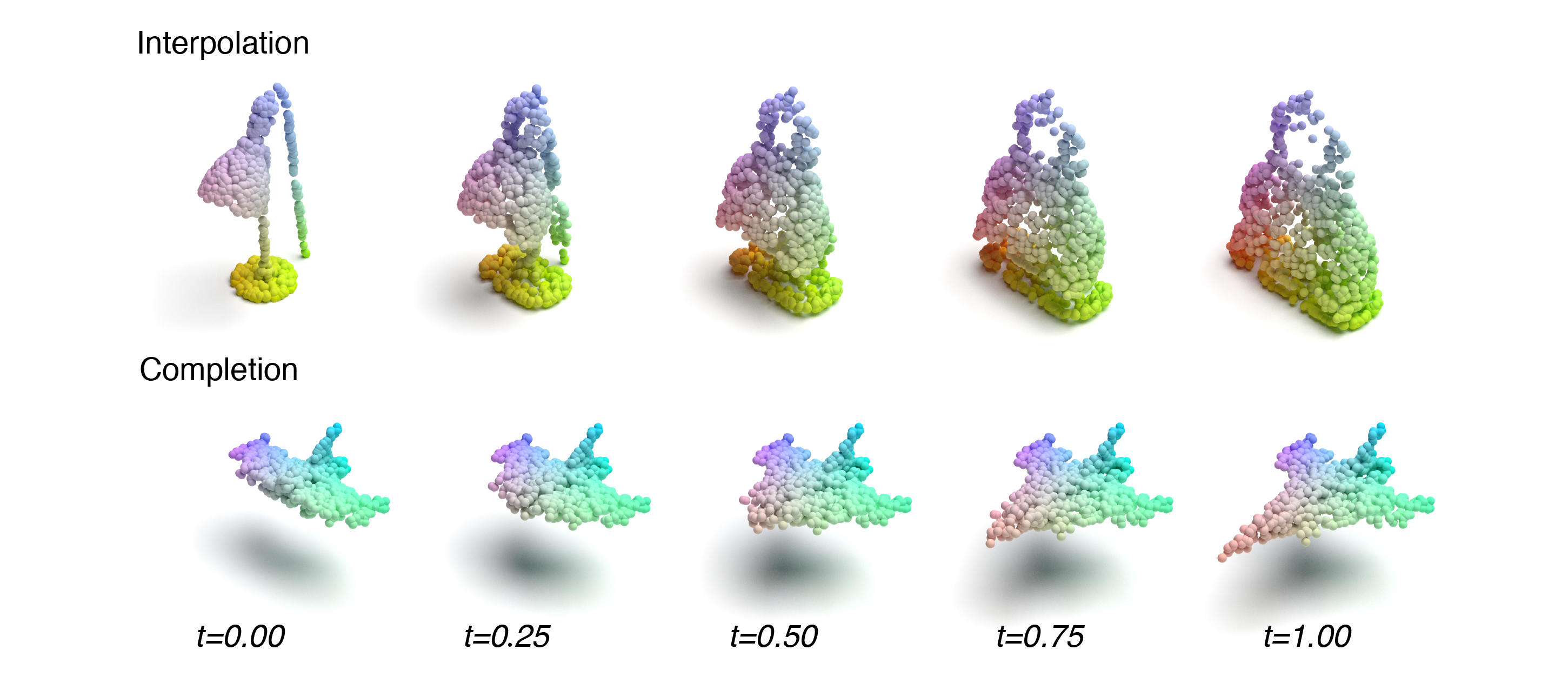}
  \vspace{-5mm}
  \caption{Interpolation and shape completion with WFM. \textit{Top}. Using the \textit{lamps} and \textit{handbags} as the source and target measures, WFM learns to transform a given (unseen test-set) lamp point-cloud into a valid handbag. \textit{Bottom}. Trained to generate full planes, WFM can reconstruct complete point-clouds from partial views of test-set samples.}\label{fig:wfm_mug2bowl}
\end{figure}

ModelNet has $40$ classes of shapes, each consisting of $n=2048$ particles. Conditioned on class label, WFM is trained to generate 
$n=1000$ sized point-clouds here too. In this setting, the noise measure is the standard normal and we did not use multi-sample matching. According to nearest-neighbour classification from OT preserving \textit{Wormhole} embeddings, generated samples match their class with an accuracy of $77.66\%$, approaching the $79.98\%$  purity of test set samples from real-data.

The MNIST dataset is a widely used collection of handwritten digits, consisting of 28x28 pixel grayscale images of the numbers 0 through 9. EMNIST (Extended MNIST) is an expansion of MNIST that includes handwritten letters as well as digits.  To convert samples from these datasets into distributions, we threshold each image and extract the coordinates of the above-threshold pixels. This produces a cohort of point-clouds of variables sizes, as each image contains a different number of relevant pixels. We apply the entropic OT map (see \cref{app:eot}) based WFM to synthesize distributions of the digit $4$ and letter $a$. Despite the data heterogeneity, WFM produces realistic examples (see \cref{fig:wfm_mnist}), while capturing the data distribution (see \cref{tab:high_d_point_clouds}).

\end{document}